\documentclass{article}

\usepackage[main, final]{neurips_2026}

\usepackage[utf8]{inputenc} 
\usepackage[T1]{fontenc}    
\usepackage{hyperref}       
\usepackage{url}            
\usepackage{graphicx}       
\usepackage{booktabs}       
\usepackage{amsmath}        
\usepackage{amsfonts}       
\usepackage{nicefrac}       
\usepackage{microtype}      
\usepackage{xcolor}         
\usepackage{tikz}           
\usetikzlibrary{
    arrows.meta,
    calc,
    fit,
    positioning,
    shapes.geometric,
    shapes.misc,
    shapes.symbols
}
\definecolor{drawioBlue}{HTML}{DAE8FC}
\definecolor{drawioBlueStroke}{HTML}{6C8EBF}
\definecolor{drawioData}{HTML}{FFF2CC}
\definecolor{drawioDataStroke}{HTML}{D6B656}
\definecolor{drawioGray}{HTML}{F5F5F5}
\definecolor{drawioGrayStroke}{HTML}{666666}
\definecolor{drawioGreen}{HTML}{D5E8D4}
\definecolor{drawioRed}{HTML}{F8CECC}
\definecolor{drawioRedStroke}{HTML}{B85450}
\definecolor{drawioTopic}{HTML}{CCFFFF}
\definecolor{drawioTopicStroke}{HTML}{0097A7}
\colorlet{drawioAnswerNode}{orange!30}
\colorlet{drawioAnswerNodeStroke}{orange!85!black}
\colorlet{drawioAnswerMark}{orange!30}
\colorlet{drawioAnswerMarkStroke}{orange!85!black}
\definecolor{drawioTextEmb}{HTML}{E1D5E7}
\definecolor{drawioTextEmbStroke}{HTML}{9673A6}
\definecolor{drawioArrow}{HTML}{666666}
\definecolor{drawioFinal}{HTML}{D5E8D4}
\definecolor{drawioFinalStroke}{HTML}{82B366}
\usepackage{tabularx}
\usepackage{multirow}
\usepackage{siunitx}
\usepackage{todonotes}
\usepackage{xspace}

\title{KG-Guard: Graph-Based Hallucination Detection for Knowledge Base Question Answering}

\author{%
  Albert Sawczyn$^{1,\dagger}$ \\
  \And Piotr Bielak$^{1}$ \And Tomasz Kajdanowicz$^{1}$ \AND
  \normalfont$^{1}$Department of Artificial Intelligence \\
  Wroc{\l}aw University of Science and Technology \\
  Wroc{\l}aw, Poland \\
  $^\dagger$\texttt{albert.sawczyn@pwr.edu.pl}\\
}

\newcommand{\methodname}{\texttt{KG-Guard}\xspace}
\begin{document}

\maketitle

\begin{abstract}
Large language models (LLMs) are increasingly used for knowledge base
question answering (KBQA), also referred to as knowledge graph question
answering (KGQA), where answering requires selecting entities
from a question-specific knowledge graph subgraph. Yet LLMs are known to
hallucinate across tasks, and KBQA is no exception: even when we provide
a graph as the knowledge source, the model may rely on
parametric knowledge instead of graph evidence or perform invalid
reasoning over the given relations.
Such hallucinated answer nodes can limit the practical
deployment of KBQA systems, especially in high-stakes domains such as
healthcare. We formulate
hallucination detection in KBQA as an answer-node classification
problem and propose a lightweight graph-based framework that treats the
answering LLM as a black box. \methodname represents each KBQA instance
as an augmented graph. It initializes node features with semantic
representations of KG entities, marks topic entities and LLM-proposed
answer nodes with learned vectors, and connects a virtual question node
to the topic entities.
A graph encoder then produces verification-oriented node representations,
and a small MLP classifies each proposed answer node using its graph
representation together with the question embedding. Experiments on
WebQSP, ComplexWebQuestions, and PUGG show that our detector achieves
the best average rank across all three benchmarks (F1 $82.0$, $87.4$,
and $84.3$), while using ${\sim}305\times$ fewer parameters than
LLM-based baselines. It also retains the best average rank when
jointly evaluated across an extended set of pipelines and settings.
Beyond
detection, the node-level feedback is actionable: when flagged answers
are fed back to the KBQA system for iterative refinement, downstream KBQA
F1 improves by $13.0$--$14.5$ points and Exact Match by
$16.9$--$17.6$ points.
\end{abstract}

\section{Introduction}
\label{sec:introduction}

Large language models (LLMs) are increasingly used as reasoning components in
knowledge-intensive applications. One important setting is
knowledge-base question answering (KBQA), also referred to as
knowledge graph question answering (KGQA),
where the model must return the entity nodes from a knowledge graph that
answer a given question \citep{lan2023complex}.
Earlier approaches queried the full graph symbolically; most LLM-based
pipelines instead retrieve a question-specific subgraph and reason over
it. For example, given ``What is the capital of Australia?'', the system
should select \emph{Canberra} from a subgraph built around the topic
entity \emph{Australia}. This graph structure provides an explicit
grounding space whose entities and relations are easier to curate,
update, and inspect than unstructured documents. Earlier
KBQA systems often relied on graph-centric architectures, whereas recent pipelines increasingly use LLMs to select
answer entities from retrieved subgraphs \citep{ma2025large, baek2023kaping,he2024gretriever}.

LLM-based pipelines improve language understanding but also introduce
hallucinations \citep{huang2025survey}. In KBQA, the LLM may rely on
parametric knowledge instead of the retrieved subgraph or reason
incorrectly over graph facts, leading to wrong answer nodes. To the best
of our knowledge, hallucination detection for LLM-based KBQA outputs
remains underexplored: prior work focuses mainly on answering KBQA or on
hallucination detection in other QA settings.

Existing detectors typically treat hallucination as a text-level problem:
they ignore the KBQA subgraph, do not classify individual answer nodes,
or require white-box access to internal LLM signals unavailable through
closed APIs. We therefore treat hallucination detection in KBQA as a graph learning
problem. Returned nodes leave structured signals in the retrieved
subgraph: hallucinated and factual nodes may differ in their connections
to topic entities and in their question-relevant local neighborhoods.

We propose \methodname, a graph-based hallucination detector for KBQA
that treats the answering LLM as a black box. Given a question, retrieved
subgraph, and LLM-proposed answer nodes, \methodname builds semantic node
and question representations, marks topic entities and answer nodes, adds
a virtual question node connected to topic entities, and runs a
lightweight graph encoder. Each returned node is classified from its
graph representation and the question embedding using a small MLP. Thus,
\methodname uses only the retrieved graph and LLM outputs, without
internal states or activations \citep{chen2024inside,binkowski2025spectral},
and avoids the extra LLM calls required by judge-based
\citep{zheng2023llmjudge} or sampling-based detectors
\citep{manakul2023selfcheckgpt}. Its node-level feedback also supports
iterative answer refinement, following evidence that fine-grained
hallucination feedback can improve factuality correction
\citep{sawczyn2025factselfcheck}.

\begin{figure*}[!htb]
    \centering
    \small
    \resizebox{0.65\textwidth}{!}{\begin{tikzpicture}[
  x=1.0cm,
  y=1.0cm,
  font=\small,
  >=Latex,
  line join=round,
  line cap=round,
  flow/.style={-Latex, thick, draw=drawioArrow},
  feedback/.style={-Latex, thick, dashed, draw=drawioRedStroke},
  data/.style={
    draw=drawioDataStroke,
    fill=drawioData,
    shape=chamfered rectangle,
    chamfered rectangle xsep=0.18cm,
    minimum width=3.0cm,
    minimum height=1.0cm,
    align=center
  },
  candidates/.style={
    draw=drawioGrayStroke,
    fill=drawioGray,
    shape=chamfered rectangle,
    chamfered rectangle xsep=0.18cm,
    minimum width=3.2cm,
    minimum height=1.0cm,
    align=center
  },
  module/.style={
    draw=drawioBlueStroke,
    rounded corners,
    fill=drawioBlue,
    minimum width=3.0cm,
    minimum height=1.0cm,
    align=center
  },
  final/.style={
    draw=drawioFinalStroke,
    fill=drawioFinal,
    shape=chamfered rectangle,
    chamfered rectangle xsep=0.18cm,
    minimum width=3.0cm,
    minimum height=1.0cm,
    align=center
  },
  flagged/.style={
    draw=drawioRedStroke,
    fill=drawioRed,
    shape=chamfered rectangle,
    chamfered rectangle xsep=0.18cm,
    minimum width=3.2cm,
    minimum height=1.0cm,
    align=center
  },
  note/.style={font=\footnotesize, align=center}
]

\node[candidates] (input) at (0.0,4.0) {KBQA instance\\$(q,G,T)$};
\node[module] (kbqa) at (4.2,4.0)
  {$f_{\mathrm{KBQA}}$\\LLM-based KBQA};
\node[candidates] (answers) at (8.4,4.0)
  {candidate answer nodes\\$\hat{A}$};
\node[module, minimum width=3.2cm, line width=0.8pt] (detector) at (8.4,2.0)
  {\textbf{\texttt{KG-Guard}}\\$f_{\theta}(q,G,T,\hat{A},\hat{a})$};
\node[flagged] (flags) at (4.2,2.0)
  {flagged set\\$\mathcal{H}=\{\hat{a}:\hat{y}_{\hat{a}}=1\}$};
\node[final] (final) at (0.0,2.0)
  {accepted answer set\\$\hat{A}$};

\draw[flow] (input) -- (kbqa);
\draw[flow] (kbqa) -- (answers);
\draw[flow] (answers) -- node[right, note] {verify each\\$\hat{a}\in\hat{A}$} (detector);
\draw[flow] (detector) -- (flags);
\draw[flow] (flags) -- node[above, note] {$\mathcal{H}=\emptyset$} (final);
\draw[feedback]
  (flags.north) -- ++(0,0.55) -| (kbqa.south);
\node[note, text=drawioRedStroke, anchor=west] at (4.30,2.95)
  {while $\mathcal{H}\neq\emptyset$};

\end{tikzpicture}}
    \caption{Role of \methodname in the KBQA loop. The LLM-based
KBQA method maps $(q,G,T)$ to candidate answer nodes $\hat{A}$.
\methodname labels returned nodes and feeds flagged hallucinations
$\mathcal{H}$ back for targeted refinement until
$\mathcal{H}=\emptyset$ or the iteration cap is reached
(see Section~\ref{sec:hallucination_correction}).}
    \label{fig:kbqa_loop}
\end{figure*}

Our contributions can be summarized as follows:
\begin{itemize}
    \item We formulate KBQA hallucination detection as answer-node
    classification on retrieved KG subgraphs -- to our knowledge, the first dedicated method for this problem.
    \item We propose \methodname, a lightweight black-box graph-based
    detector that achieves the best average rank across all three
    benchmarks, using ${\sim}305\times$ fewer parameters than
    LLM-based baselines. It also retains the best average rank when
    jointly evaluated across an extended set of pipelines and settings.
    \item We show that node-level feedback enables iterative refinement
    (Fig.~\ref{fig:kbqa_loop}), improving KBQA F1 by $13.0$--$14.5$
    points and Exact Match by $16.9$--$17.6$ points.
    \item We evaluate on WebQSP, CWQ, and PUGG against LLM-judge,
    sampling-based, and fine-tuned text-encoder baselines, ablate key
    design choices, and test robustness across answering LLMs,
    retrieval methods, and KBQA pipelines.
    \item We release all code, intermediate data, and a KBQA
    hallucination detection dataset.
    \footnote{Code: 
    \href{https://github.com/graphml-lab-pwr/KG-Guard}{github.com/graphml-lab-pwr/KG-Guard} (License: CC BY-SA 4.0);
    Dataset: \href{https://huggingface.co/collections/graphml-lab-pwr/kg-guard-datasets}{huggingface.co/collections/graphml-lab-pwr/kg-guard-datasets}
    (License: CC BY-SA 4.0 (WebQSP and PUGG subsets); GPL v3 (CWQ subset)).}
\end{itemize}

\section{Related Work}
\label{sec:related}

Recent KBQA systems increasingly combine language models with retrieved
graph evidence. KAPING augments prompts with relevant KG facts for
zero-shot question answering \citep{baek2023kaping}. Earlier graph QA
models, such as EmbedKGQA, GRAFT-Net, QA-GNN, and GreaseLM, rank
candidates via KG embeddings, question-aware subgraph reasoning, or
joint language--graph representations, without an LLM generator
\citep{saxena2020embedkgqa,sun2018graftnet,yasunaga2021qagnn,zhang2022greaselm}.
A related approach, G-Retriever, targets open-ended QA over
KG, applying retrieval-augmented generation (RAG) to
construct query-relevant subgraphs via prize-collecting Steiner tree
(PCST) optimization
\citep{he2024gretriever}. More recent agentic approaches traverse or
plan over KG reasoning paths iteratively to produce grounded answers
\citep{sun2024thinkongraph,luo2024reasoningongraphs}. GNN-RAG feeds 
GNN-retrieved KG reasoning paths to an LLM for answer generation \citep{mavromatis2024gnnrag}. 
These methods use graph evidence for question answering. 
Our goal is orthogonal: given candidate answer nodes from an external
LLM-based KBQA system, we ask whether they are hallucinated -- a problem,
to our knowledge, not previously studied.

Hallucination detection has largely been studied in free-text settings.
Black-box sampling-based methods such as SelfCheckGPT estimate factuality by 
sampling multiple generations \citep{manakul2023selfcheckgpt}.
Fine-grained methods verify outputs at the level of individual facts or
claims \citep{min2023factscore,sawczyn2025factselfcheck}; our work
shares this spirit by classifying individual answer nodes rather than
whole generations. Another
line of work exploits internal model signals such as hidden states
\citep{azaria2023internal,chen2024inside,kossen2024semanticentropyprobes,farquhar2024detecting}
or attention maps
\citep{chuang2024lookback,sriramanan2024llmcheck,binkowski2025spectral}.
These methods use internal LLM signals to detect hallucinations, but
they are not designed to verify answer nodes against a retrieved KBQA
subgraph. Their access requirements also limit applicability to closed
models.

Several recent approaches apply the KG structure to hallucination detection. 
GraphEval \citep{sansford2024grapheval} extracts atomic claims from LLM 
output as KG triples and verifies each against the provided textual context 
via natural language inference (NLI) models; FactAlign \citep{rashad2024factalign} and knowledge-centric
detection \citep{hu2024knowledge} similarly extract triples from generated 
text and align them with a textual reference. GraphCheck constructs KGs from 
both the claim and source document, then applies GNNs as a soft prompt to an 
LLM verifier \citep{chen2025graphcheck}. These approaches verify
free-form generated text using external textual evidence, whereas we classify
answer nodes directly on an existing structured KG.

Iterative refinement is a general strategy for correcting LLM outputs
\citep{madaan2023selfrefine,dhuliawala2024cove, sawczyn2025factselfcheck}. 
KGR uses direct triple lookup in a KG to guide revision \citep{guan2024kgr}. Our
correction procedure identifies hallucinated answer nodes
via a trained graph-encoder classifier rather than triple-level conflict
resolution.

Message-passing GNNs compute node representations by repeatedly
aggregating information from local neighborhoods, as in GCN, GraphSAGE,
and GIN \citep{kipf2017semi,hamilton2017inductive,xu2019powerful}, with
GINE extending GIN to incorporate edge features
\citep{hu2020strategies}.
For our task, attention-based graph
encoders are especially natural because only part of the retrieved graph
may be relevant. Graph Attention Networks (GAT)
learn neighbor-specific attention weights \citep{velivckovic2018gat},
while GraphTransformer extends this with multi-head dot-product attention for expressive message passing
\citep{dwivedi2021generalization,shi2021maskedlabel}.

\section{Method}
\label{sec:method}

\subsection{Problem Formulation}
\label{sec:problem_formulation}

We consider a KBQA task in which each example is associated with a
natural-language question and a question-specific subgraph extracted
from a knowledge graph. Formally, each instance is represented as
$(q, G, T, A^{*})$, where $q$ is the question and $G = (V, E)$ is the
retrieved subgraph. Here, $V$ is the set of nodes and each edge
$e=(u,r,v) \in E$ corresponds to a directed KG triple from node $u$ to
node $v$ with relation label $r \in \mathcal{R}$, where
$\mathcal{R}$ denotes the set of relation labels. We treat these
relations as edge attributes rather than as a fixed heterogeneous graph
schema, which allows relation labels to be encoded from their text. The
set $T \subseteq V$ contains the topic entities, i.e., nodes
mentioned in the question. Let $V_{\mathrm{KG}}$ denote the full KG entity set;
$A^{*} \subseteq V_{\mathrm{KG}}$ contains the gold answer nodes,
which need not all occur in the retrieved node set $V$.

Given $(q, G, T)$, an LLM-based KBQA method
$f_{\mathrm{KBQA}}$ is applied to answer the question:
\[
    \hat{A} = f_{\mathrm{KBQA}}(q, G, T), \qquad \hat{A} \subseteq V .
\]
If the method abstains with a special \emph{unknown} response, we set
$\hat{A} = \emptyset$. In this work, we only consider examples for which
the method returns at least one node, i.e., $\hat{A} \neq \emptyset$.
This restriction matches the goal of our detector, which is to verify
concrete answer nodes rather than abstentions.

The hallucination detection task is then formulated as a binary
classification problem over individual returned nodes. If the LLM
returns multiple nodes for a single question, we decompose this output
into separate detection instances, one per predicted node. Thus, for
each retained KBQA example $(q, G, T, A^{*}, \hat{A})$ and each
$\hat{a} \in \hat{A}$, we create one classification instance
$(q, G, T, \hat{A}, A^{*}, \hat{a})$ and define the target label
$y \in \{0, 1\}$ as
\[
    y =
    \begin{cases}
        0, & \text{if } \hat{a} \in A^{*}, \\
        1, & \text{otherwise,}
    \end{cases}
\]
where $y = 1$ denotes a hallucinated answer node and $y = 0$ denotes a
factual one. This definition naturally handles questions with multiple
valid answers: a returned node is treated as factual if it matches any
gold answer node, and hallucinated otherwise. Consequently, a
single KBQA example may yield multiple hallucination-detection
instances, corresponding to the different nodes proposed by the LLM. 
In a deployed system, however, the detector can process the graph once 
(single forward pass) and mark which of the returned nodes are hallucinated. 

Our goal is to learn a detector
$f_{\theta}(q, G, T, \hat{A}, \hat{a}) \in \{0, 1\}$ that predicts whether
an individual node selected by $f_{\mathrm{KBQA}}$ is factual or hallucinated,
where $\theta$ collectively denotes all trainable parameters of the detector.

\subsection{\texorpdfstring{\texttt{KG-Guard}}{KG-Guard} (Hallucination Detector)}
\label{sec:hallucination_detector}

\begin{figure*}[!htb]
    \centering
    \small
    \resizebox{0.90\textwidth}{!}{\begin{tikzpicture}[
  x=1.0cm,
  y=1.0cm,
  font=\small,
  >=Latex,
  line join=round,
  line cap=round,
  flow/.style={-Latex, thick, draw=drawioArrow},
  kgflow/.style={-Latex, thick, draw=drawioArrow},
  qflow/.style={-Latex, thick, dashed, draw=drawioDataStroke},
  module/.style={
    draw=drawioBlueStroke,
    rounded corners,
    fill=drawioBlue,
    minimum width=2.4cm,
    minimum height=1.0cm,
    align=center
  },
  data/.style={
    draw=drawioDataStroke,
    fill=drawioData,
    shape=chamfered rectangle,
    chamfered rectangle xsep=0.18cm,
    minimum width=1.75cm,
    minimum height=0.85cm,
    align=center
  },
  output/.style={
    draw=drawioGrayStroke,
    fill=white,
    shape=chamfered rectangle,
    chamfered rectangle xsep=0.18cm,
    minimum width=2.75cm,
    minimum height=1.0cm,
    inner xsep=2pt,
    inner ysep=2pt,
    align=center,
    path picture={
      \fill[drawioRed]
        (path picture bounding box.north west) rectangle
        (path picture bounding box.east);
      \fill[drawioGreen]
        (path picture bounding box.west) rectangle
        (path picture bounding box.south east);
      \draw[drawioGrayStroke, opacity=0.45]
        (path picture bounding box.west) --
        (path picture bounding box.east);
    }
  },
  eqtextpart/.style={
    draw=drawioTextEmbStroke,
    fill=drawioTextEmb,
    rounded corners=1pt,
    inner xsep=2pt,
    inner ysep=1pt,
    minimum height=0.36cm,
    font=\scriptsize
  },
  eqtopicpart/.style={
    draw=drawioGrayStroke,
    rounded corners=1pt,
    inner xsep=2pt,
    inner ysep=1pt,
    minimum width=1.1cm,
    minimum height=0.36cm,
    font=\scriptsize,
    path picture={
      \fill[drawioGray, fill opacity=0.28]
        (path picture bounding box.north west) --
        (path picture bounding box.north east) --
        (path picture bounding box.south east) -- cycle;
      \fill[drawioTopic, fill opacity=0.75]
        (path picture bounding box.north west) --
        (path picture bounding box.south west) --
        (path picture bounding box.south east) -- cycle;
      \draw[drawioGrayStroke, opacity=0.45]
        (path picture bounding box.north west) --
        (path picture bounding box.south east);
    }
  },
  eqanswerpart/.style={
    draw=drawioGrayStroke,
    rounded corners=1pt,
    inner xsep=2pt,
    inner ysep=1pt,
    minimum width=1.1cm,
    minimum height=0.36cm,
    font=\scriptsize,
    path picture={
      \fill[drawioGray, fill opacity=0.28]
        (path picture bounding box.north west) --
        (path picture bounding box.north east) --
        (path picture bounding box.south east) -- cycle;
      \fill[drawioAnswerMark, fill opacity=0.75]
        (path picture bounding box.north west) --
        (path picture bounding box.south west) --
        (path picture bounding box.south east) -- cycle;
      \draw[drawioGrayStroke, opacity=0.45]
        (path picture bounding box.north west) --
        (path picture bounding box.south east);
    }
  },
  eqvqpart/.style={
    draw=drawioGrayStroke,
    rounded corners=1pt,
    inner xsep=2pt,
    inner ysep=1pt,
    minimum width=0.5cm,
    minimum height=0.36cm,
    font=\scriptsize,
    path picture={
      \fill[drawioGray, fill opacity=0.28]
        (path picture bounding box.north west) --
        (path picture bounding box.north east) --
        (path picture bounding box.south east) -- cycle;
      \fill[drawioData, fill opacity=0.75]
        (path picture bounding box.north west) --
        (path picture bounding box.south west) --
        (path picture bounding box.south east) -- cycle;
      \draw[drawioGrayStroke, opacity=0.45]
        (path picture bounding box.north west) --
        (path picture bounding box.south east);
    }
  },
  note/.style={font=\footnotesize, align=center},
  qnode/.style={
    circle,
    draw=drawioTopicStroke,
    fill=drawioTopic,
    minimum size=5.2mm,
    inner sep=0pt
  },
  vqnode/.style={
    circle,
    draw=drawioDataStroke,
    fill=drawioData,
    minimum size=5.2mm,
    inner sep=0pt
  },
  anode/.style={
    circle,
    draw=drawioAnswerNodeStroke,
    fill=drawioAnswerNode,
    minimum size=5.2mm,
    inner sep=0pt
  },
  bothnode/.style={
    circle,
    draw=drawioAnswerNodeStroke,
    fill=drawioTopic,
    double=drawioAnswerNodeStroke,
    double distance=1pt,
    minimum size=5.2mm,
    inner sep=0pt
  },
  gnode/.style={
    circle,
    draw=drawioGrayStroke,
    fill=drawioGray,
    minimum size=5.2mm,
    inner sep=0pt
  }
]

\node[note] at (2.55,5.20) {Augmented graph $\widetilde{G}$};
\node[vqnode] (vq) at (0.95,4.60) {$v_q$};
\node[qnode] (v1) at (1.75,3.60) {};
\node[qnode] (v2) at (2.95,4.35) {};
\node[anode] (v3) at (4.20,3.60) {};
\node[gnode] (v4) at (2.75,2.55) {};
\node[gnode] (v5) at (4.20,2.35) {};
\node[anode] (v6) at (3.15,1.75) {};
\draw[qflow] (vq) to[bend left=10] (v1);
\draw[qflow] (v1) to[bend left=10] (vq);
\draw[qflow] (vq) to[bend left=10] (v2);
\draw[qflow] (v2) to[bend left=10] (vq);
\draw[kgflow] (v1) -- (v2);
\draw[kgflow] (v2) -- (v3);
\draw[kgflow] (v1) -- (v4);
\draw[kgflow] (v4) -- (v3);
\draw[kgflow] (v3) -- (v5);
\draw[kgflow] (v4) -- (v6);
\draw[kgflow] (v6) -- (v5);

\begin{scope}[shift={(-0.70,4.04)}]
  \foreach \i in {0,...,2} {
    \filldraw[draw=drawioGrayStroke, fill=drawioData]
      (\i*0.22,0) rectangle +(0.18,0.32);
  }
  \filldraw[draw=drawioGrayStroke, fill=drawioGray]
    (0.69,0) rectangle +(0.18,0.32);
  \filldraw[draw=drawioGrayStroke, fill=drawioGray]
    (0.91,0) rectangle +(0.18,0.32);
  \filldraw[draw=drawioDataStroke, fill=drawioData]
    (1.13,0) rectangle +(0.18,0.32);
\end{scope}

\begin{scope}[shift={(0.20,3.05)}]
  \foreach \i in {0,...,2} {
    \filldraw[draw=drawioTextEmbStroke, fill=drawioTextEmb]
      (\i*0.22,0) rectangle +(0.18,0.32);
  }
  \filldraw[draw=drawioTopicStroke, fill=drawioTopic]
    (0.69,0) rectangle +(0.18,0.32);
  \filldraw[draw=drawioGrayStroke, fill=drawioGray]
    (0.91,0) rectangle +(0.18,0.32);
  \filldraw[draw=drawioGrayStroke, fill=drawioGray]
    (1.13,0) rectangle +(0.18,0.32);
\end{scope}

\begin{scope}[shift={(3.85,4.00)}]
  \foreach \i in {0,...,2} {
    \filldraw[draw=drawioTextEmbStroke, fill=drawioTextEmb]
      (\i*0.22,0) rectangle +(0.18,0.32);
  }
  \filldraw[draw=drawioGrayStroke, fill=drawioGray]
    (0.69,0) rectangle +(0.18,0.32);
  \filldraw[draw=drawioAnswerMarkStroke, fill=drawioAnswerMark]
    (0.91,0) rectangle +(0.18,0.32);
  \filldraw[draw=drawioGrayStroke, fill=drawioGray]
    (1.13,0) rectangle +(0.18,0.32);
\end{scope}

\begin{scope}[shift={(1.15,2.05)}]
  \foreach \i in {0,...,2} {
    \filldraw[draw=drawioTextEmbStroke, fill=drawioTextEmb]
      (\i*0.22,0) rectangle +(0.18,0.32);
  }
  \filldraw[draw=drawioGrayStroke, fill=drawioGray]
    (0.69,0) rectangle +(0.18,0.32);
  \filldraw[draw=drawioGrayStroke, fill=drawioGray]
    (0.91,0) rectangle +(0.18,0.32);
  \filldraw[draw=drawioGrayStroke, fill=drawioGray]
    (1.13,0) rectangle +(0.18,0.32);
\end{scope}

\node[note, anchor=west] (eqstart) at (-0.25,1.25) {$x_v = [$};
\node[eqtextpart, right=-0.02cm of eqstart] (eqphi) {$\phi(t_v)$};
\node[note, right=-0.02cm of eqphi] (eqsepone) {$\|$};
\node[eqtopicpart, right=-0.02cm of eqsepone] (eqtopic) {$M_T[\tau_v]$};
\node[note, right=-0.02cm of eqtopic] (eqseptwo) {$\|$};
\node[eqanswerpart, right=-0.02cm of eqseptwo] (eqanswer) {$M_A[\alpha_v]$};
\node[note, right=-0.02cm of eqanswer] (eqsepthree) {$\|$};
\node[eqvqpart, right=-0.02cm of eqsepthree] (eqvq) {$\mu_v$};
\node[note, right=-0.02cm of eqvq] (eqend) {$]$};
\node[note, text width=4.6cm, align=left] at (2.15,0.45)
  {\textcolor{drawioDataStroke}{$v_q$}: virtual question node\\
  \textcolor{drawioTopicStroke}{cyan}: topic entities\\
  \textcolor{drawioAnswerMarkStroke}{orange}: answer nodes};

\node[module] (genc) at (7.10,3.55) {graph encoder\\$g_{\theta}(\cdot)$};
\draw[flow] (4.95,3.55) -- (genc);

\begin{scope}[shift={(9.47,3.35)}]
  \foreach \i in {0,...,2} {
    \filldraw[draw=drawioAnswerNodeStroke, fill=drawioAnswerNode]
      (\i*0.22,0) rectangle +(0.18,0.40);
  }
\end{scope}
\node[note] at (9.80,4.25) {answer-node\\embedding $h_{\hat{a}}$};
\draw[flow] (genc.east) -- (9.37,3.55);

\node[
  data,
  minimum width=1.25cm,
  inner xsep=2pt,
  inner ysep=1pt
] (qtext) at (6.50,0.65) {question\\$q$};
\node[module, minimum width=1.75cm] (phi) at (8.75,0.65)
  {text encoder\\$\phi(\cdot)$};
\draw[flow] (qtext) -- (phi);

\begin{scope}[shift={(10.13,0.45)}]
  \foreach \i in {0,...,2} {
    \filldraw[draw=drawioGrayStroke, fill=drawioData]
      (\i*0.22,0) rectangle +(0.18,0.40);
  }
\end{scope}
\node[note] at (10.46,0.05) {question\\embedding $z_q$};
\draw[flow] (phi.east) -- (10.03,0.65);

\begin{scope}[shift={(9.47,2.10)}]
  \foreach \i in {0,...,2} {
    \filldraw[draw=drawioAnswerNodeStroke, fill=drawioAnswerNode]
      (\i*0.22,0) rectangle +(0.18,0.40);
  }
  \foreach \i in {3,...,5} {
    \filldraw[draw=drawioGrayStroke, fill=drawioData]
      (\i*0.22,0) rectangle +(0.18,0.40);
  }
\end{scope}
\node[note, anchor=east] at (9.22,2.30) {$[h_{\hat{a}}\,\|\,z_q]$};
\draw[flow] (9.78,3.35) -- (9.78,2.50);
\draw[flow] (10.44,0.85) -- (10.44,2.10);

\node[
  module,
  minimum width=1.10cm,
  inner xsep=2pt,
  inner ysep=1pt
] (mlp) at (11.75,2.30)
  {MLP\\$\psi(\cdot)$};
\node[output] (score) at (14.55,2.30)
  {Hallucinated ($y=1$)\\or factual ($y=0$)};
\draw[flow] (10.85,2.30) -- (mlp);
\draw[flow] (mlp) -- (score);

\end{tikzpicture}}
    \caption{\methodname architecture for classifying answer nodes.
    Node features combine semantic node representations with
    topic-entity marks $M_T$ and answer-node marks $M_A$. A virtual
    question node $v_q$ is connected to the topic entities with edges in
    both directions. The graph encoder $g_{\theta}$ computes answer-node
    representations $h_{\hat{a}}$, which are concatenated with the
    question embedding $z_q$ and passed to an MLP to predict whether the
    returned node is hallucinated ($y=1$) or factual ($y=0$).}
    \label{fig:detector_pipeline}
\end{figure*}

Our method (see Figure~\ref{fig:detector_pipeline}) operates on top of
the answer returned by the LLM-based KBQA method for a given KBQA
instance (assuming $\hat{A}\neq\emptyset$).

We start from the retrieved subgraph and construct node features by
combining semantic node information with task-specific marks.
Knowledge-graph nodes typically do not come with informative numeric
features, so we initialize them from text. Let $t_v$ denote the textual
representation of node $v$, such as its label or name, and let
$\phi(\cdot)$ be a semantic text encoder. For each KG edge $(u,r,v)$,
the same encoder maps the relation label $t_r$ to the edge feature
$\phi(t_r)$.

We then augment these semantic features with trainable mark embeddings.
Let $\tau_v = \mathbb{I}[v \in T]$ indicate whether node $v$ is a topic
entity, and let $\alpha_v = \mathbb{I}[v \in \hat{A}]$ indicate whether
it was returned by the LLM-based KBQA method. We use two trainable lookup
tables, $M_T \in \mathbb{R}^{2 \times d_T}$ and
$M_A \in \mathbb{R}^{2 \times d_A}$, for topic-entity and answer-node
marks, respectively. Each table contains one embedding for the positive
case and one embedding for the negative case.

The initial feature vector of node $v$ is then:
\[
    x_v =
    \left[
        \phi(t_v)
        \,\|\, M_T[\tau_v]
        \,\|\, M_A[\alpha_v]
        \,\|\, \mu_v
    \right],
\]
where $\|$ denotes concatenation and $\mu_v = \mathbb{I}[v = v_q]$ flags
the virtual question node introduced below. Thus, each node
representation contains four parts: a semantic representation of the
node text, a learnable mark indicating whether the node is a topic
entity, a learnable mark indicating whether the node was proposed as an
answer, and a flag for the virtual question node. We use the same text
encoder to obtain the question embedding $z_q = \phi(q)$.

To condition message passing on the question, we augment the graph with
a virtual question node $v_q$, built from the same feature template
$x_v$ whose semantic part is $z_q$ instead of $\phi(t_v)$. This node is
connected only to the topic entities $T$, in both directions, rather
than to every node in the retrieved subgraph. These edges have no
relation label, so each direction uses a learnable edge embedding. Let
$\widetilde{G}$ and $\widetilde{X}$ denote the resulting augmented
graph and its node features.

The marked and augmented graph is then processed by a graph encoder
$g_{\theta}$ to obtain node-level representations
$H = g_{\theta}(\widetilde{G}, \widetilde{X})$, where
$H = \{h_v\}_{v \in V}$. We instantiate $g_{\theta}$ as a
GraphTransformer, GAT, or GINE, with the GraphTransformer used as the main
configuration in our experiments. For a particular returned answer node
$\hat{a}$, we extract its final representation $h_{\hat{a}}$ and
concatenate it with the question embedding $z_q$. The resulting vector
is passed through an MLP classifier, with parameters $\psi \subset \theta$,
to obtain a scalar logit:
\[
    s_{\hat{a}} =
    \mathrm{MLP}_{\psi}\left([h_{\hat{a}} \,\|\, z_q]\right),
\]
where $s_{\hat{a}}$ is a scalar logit. The final hallucination score is
obtained as
\[
    \hat{p}(y = 1 \mid q, G, T, \hat{A}, \hat{a}) = \sigma(s_{\hat{a}}),
\]
with $y = 1$ denoting a hallucinated answer node. Thus, the detector
combines graph structure around the returned node with the semantics of
the question, while remaining independent of the internal activations
of the answering LLM.

\section{Experimental Setup}
\label{sec:experiments}

The extended implementation details are provided in
Appendix~\ref{sec:implementation_details}.

\subsection{Evaluation Data}
\label{sec:experiments-datasets}

Training and evaluation of our detector require tuples
$(q, G, T, A^{*}, \hat{A})$ containing a question, a retrieved subgraph,
topic entities, gold answer nodes, and LLM-predicted answer nodes.
The KBQA benchmarks used here do not directly provide these complete tuples. Therefore, we
construct the evaluation data by running a full KBQA
pipeline on three established benchmarks, while recording both gold and
predicted answer nodes as detection instances.
In practical deployment, directly annotating the correctness of
proposed answer nodes is another way to obtain detector labels.

\subsubsection{KBQA Benchmarks}
\label{sec:experiments-datasets-kbqa}

We evaluate on three KBQA benchmarks that vary in question complexity,
underlying knowledge graph, and language. We inherit each benchmark's
official question-level splits, assigning every answer-node instance
from a question to the same
split to prevent leakage between training and evaluation.
\paragraph{WebQuestionsSP} (\mbox{WebQSP}) \citep{yih2016webqsp} (license: CC BY 4.0) is a
standard factoid KBQA benchmark grounded in the Freebase knowledge graph \citep{bollacker2007freebase}. Questions
are naturally phrased real-world queries collected from Google Suggest
\citep{berant2013webq}. Because the validation split provided with this
dataset is small (246 questions), we supplement it with questions
sampled from the training set to reach a validation set size of 500
questions.

\paragraph{ComplexWebQuestions} (\mbox{CWQ}) \citep{talmor2018complex} (license: GPL v2+)
extends WebQSP by programmatically appending compositional constraints to
WebQSP SPARQL queries, then paraphrasing the resulting questions via
crowdworkers. This procedure produces questions requiring logical reasoning,
making CWQ substantially harder than
WebQSP. It is also grounded in Freebase and is the largest of our three
benchmarks.

\paragraph{PUGG} \citep{sawczyn2024pugg} (license: CC BY-SA 4.0) is a Polish KBQA dataset grounded
in the Wikidata knowledge graph \citep{vrandevcic2014wikidata}, combining naturally collected (Google Suggest) and
template-based (SPARQL-paired templates and paraphrasing) questions.
We use PUGG to evaluate KBQA on a non-English language and a different KG.
Because Wikidata hub nodes inflate subgraph sizes, we exclude nodes
connected to more than $1000$ other entities as a preprocessing step before subgraph retrieval.

\subsubsection{Hallucination Dataset Construction}
\label{sec:experiments-datasets-hallu}

\begin{table}[!htb]
\small
\caption{Dataset statistics for hallucination detection. Each KBQA question may yield multiple proposed answer nodes. \underline{\#Answers} counts these nodes plus abstention entries. \underline{\%Hallu.}, \underline{\%Correct}, and \underline{\%Abstained} use this combined count as their denominator. Abstention entries are excluded from node classification.}
\label{tab:dataset_overview}
\centering
\begin{tabular}{llrrrrr}
\toprule
\textbf{Dataset} & \textbf{Split} & \textbf{\#Questions} & \textbf{\#Answers} & \textbf{\%Hallu.} & \textbf{\%Correct} & \textbf{\%Abstained} \\
\midrule
\multirow{3}{*}{\textbf{WebQSP}} & train & 2826 & 9179 & 50.0 & 44.5 & 5.5 \\
 & val & 246 & 807 & 54.0 & 41.0 & 5.0 \\
 & test & 1628 & 4842 & 46.2 & 47.9 & 5.8 \\
\midrule
\multirow{3}{*}{\textbf{CWQ}} & train & 27639 & 43795 & 47.6 & 28.3 & 24.2 \\
 & val & 3519 & 6001 & 55.1 & 24.1 & 20.8 \\
 & test & 3531 & 5840 & 55.1 & 21.7 & 23.1 \\
\midrule
\multirow{3}{*}{\textbf{PUGG}} & train & 3589 & 6372 & 63.6 & 24.3 & 12.1 \\
 & val & 700 & 1078 & 58.6 & 28.9 & 12.4 \\
 & test & 1081 & 1884 & 62.9 & 24.9 & 12.2 \\
\bottomrule
\end{tabular}
\end{table}

\paragraph{Subgraph retrieval} Both KBQA and hallucination detection require
a question-relevant subgraph from the KG; we apply the same retrieval
procedure for both tasks. For all three datasets, we follow the
\mbox{G-Retriever} pipeline \citep{he2024gretriever}: candidate neighbors
are first collected by BFS expansion from topic entities, and a
Prize-Collecting Steiner Tree (PCST) formulation is then solved to select a
compact, question-relevant subgraph. We use the same PCST
hyperparameters as in the original \mbox{G-Retriever} pipeline. Subgraph
statistics are reported in
Appendix~\ref{sec:graph_stats} (Table~\ref{tab:graph_stats}).

As an alternative retrieval strategy, we also evaluate top-$K{=}100$
embedding retrieval to assess the robustness of \methodname to the
choice of subgraph construction method; details and results are in
Appendix~\ref{sec:additional_results-topk}.

\paragraph{KBQA method} We use KAPING \citep{baek2023kaping}
as the LLM-based KBQA method. Beyond the answer node, KAPING is extended to also return a textual reasoning summary and a list of reasoning triples (the KG triples
cited as evidence). Our approach does not use these elements, and they are
used only as additional context for the LLM-based baselines
(Section~\ref{sec:baselines}). The answering LLM is constrained to return
only nodes present in the retrieved subgraph via structured generation; it
may also abstain with an \emph{unknown} response. The full prompt
template is provided in the code repository. As discussed in
Section~\ref{sec:problem_formulation}, abstaining examples are excluded
from the hallucination-detection task. The statistics, including the fraction of unknown
responses, are reported in Table~\ref{tab:dataset_overview}. 
Additional statistics of KBQA answers are reported in 
Appendices~\ref{sec:answer_stats} and~\ref{sec:abstention_patterns}.

In addition to KAPING, we evaluate \methodname on the
\mbox{G-Retriever} pipeline \citep{he2024gretriever}, which uses the
same PCST subgraph retrieval and a fine-tuned LLM to select answer
nodes from the subgraph, to assess robustness to the choice of KBQA
pipeline. Since its answering LLM returns free text rather than
structured node identifiers, we parse \mbox{G-Retriever}'s output
into a set of answer nodes using \mbox{GPT-5.4-nano} as an extraction
LLM. Results are reported in
Appendix~\ref{sec:additional_results-gretriever}. Hallucination labels
are specific to the answering pipeline, generator, and retriever, so
we train and evaluate a separate \methodname model for each
configuration rather than reusing a detector in a zero-shot manner.

\subsection{Baselines}
\label{sec:baselines}

We compare against four groups of baselines covering trivial predictors,
a fine-tuned text encoder, LLM-based detection, and sampling-based
consistency analysis. All baselines operate without access to the
answering model's internals, mirroring the common real-world deployment
setting where the model is accessed as a service.

\paragraph{Trivial baselines} \emph{Random} assigns the hallucinated
label uniformly at random; \emph{Most Frequent} always predicts the
majority class in the validation split. These provide reference points
and illustrate the effect of class imbalance on each metric.

\paragraph{Fine-tuned text encoder}
We fine-tune \mbox{RoBERTa-large} \citep{liu2019roberta} separately on
WebQSP and CWQ, and \mbox{polish-RoBERTa-large-v2} \citep{dadas2020pre}
on PUGG. Unlike the prompted LLM judges, it is a task-specific
discriminative detector trained on the same per-answer-node labels as
\methodname.
Each example concatenates the question, the predicted answer, and a
subgraph serialized as a triple list. Unlike \methodname, it treats the
retrieved graph as a flat text sequence rather than performing message
passing over its nodes and edges.

\paragraph{Llama-4-Scout} We use \mbox{Llama-4-Scout}
\citep{meta2025llama4} as an LLM-as-judge baseline \citep{zheng2023llmjudge,janiak2025illusion}.
We evaluate three prompt variants supplying progressively richer context:
\textbf{(1)} \emph{Graph} (G): question, subgraph serialized as a
triple list, and the predicted answers to verify;
\textbf{(2)} \emph{Graph + Reasoning Summary} (G+RS): the above
plus the answering LLM's reasoning summary;
\textbf{(3)} \emph{Graph + Reasoning Summary + Reasoning Triples}
(G+RS+RT): the above plus the answering LLM's reasoning triples.
While straightforward to deploy, this approach is computationally
expensive: \mbox{Llama-4-Scout} has $109$B total parameters ($17$B
active in its MoE architecture), requiring substantial GPU memory per
detection call.

\paragraph{GPT-5-mini} We evaluate \mbox{GPT-5-mini}
\citep{openai2025gpt5mini} as a judge using two configurations:
G and G+RS+RT. Both outperformed G+RS on average with
\mbox{Llama-4-Scout}. Due to commercial API costs, we restrict evaluation to these
two configurations.

\paragraph{SelfCheckGPT} We adapt the sampling-based method
of \citet{manakul2023selfcheckgpt} to structured KBQA outputs. We draw
$N$ independent predictions from the LLM-based KBQA method with
temperature $1.0$. We then compute the fraction of predictions that do
\emph{not} include a given node as the hallucination score. We evaluate
with $N \in \{5, 10\}$ to assess sensitivity to sample count. This
baseline represents a widely used black-box hallucination detection
method and is the most expensive computationally, requiring $N$ forward
passes of the LLM-based KBQA method per example.

\subsection{Evaluation} 

For hallucination detection, we report F1 as the primary metric, computed over
predicted answer nodes only; Precision, Recall, Accuracy, and AUC-PR are
reported in Appendix~\ref{sec:additional_results-llama_full}.
Trainable detectors are trained with $3$ random seeds; we report the
mean and standard deviation across seeds. We also report Random over
$3$ seeds. For methods that rely on LLM calls, we report single-run
results due to the high cost of inference.

\subsection{Hallucination Correction}
\label{sec:hallucination_correction}

While our primary focus is hallucination detection, we also examine
its potential for answer correction. \methodname can guide answer
revision within a larger system (Fig.~\ref{fig:kbqa_loop}). We
investigate whether feeding
\methodname predictions back to the LLM enables it to revise its own
hallucinated responses.

Starting from the initial \mbox{KAPING} answers,
we iterate the following steps: \textbf{(1)}~ \methodname scores each 
predicted answer node. \textbf{(2)}~Examples with no nodes flagged as 
hallucinated are marked resolved, and their answers remain unchanged. 
\textbf{(3)}~For each active example (i.e., with at least one flagged 
answer node), the LLM is re-prompted in a chat-style conversation consisting
of a system instruction, the original question and subgraph, the LLM's most recent
response, and a follow-up message listing the flagged answer node
names. Steps~1--3 repeat until all examples are resolved or
the iteration cap (max. $5$) is reached. We use \mbox{Llama-4-Scout}
as the refinement LLM.

We report downstream KBQA metrics (F1, Precision, Recall,
and Exact Match) at the initial step and after refinement on all three
datasets, together with the distribution of refinement iterations.

\section{Results}
\label{sec:results}

Ablation studies on graph encoder components, input modalities,
virtual question-node connectivity, and graph encoder architecture,
together with training on KBQA-derived labels and calibration, are
presented in Appendix~\ref{sec:additional_experiments}.

\subsection{Hallucination Detection}
\label{sec:results-hallucination_detection}

Hallucination detection results with \mbox{Llama-4-Scout} as the
answering LLM are summarized in Table~\ref{tab:results_summary}.
Full per-metric results are in
Appendix~\ref{sec:additional_results-llama_full}.

\methodname (GraphTransformer) has the best average rank ($1.7$) and
achieves the highest F1 on WebQSP ($82.0$). On CWQ, it reaches
$87.4$, $0.3$ points below the fine-tuned RoBERTa baseline; on PUGG,
it reaches $84.3$, $1.2$ points below SelfCheckGPT ($N{=}10$,
$85.5$). SelfCheckGPT ($N{=}10$) is the strongest baseline on average, 
but requires $10$ generations from the answering
LLM per example, which is extremely expensive, versus a
single feed-forward pass for \methodname.

\begin{table}[!htb]
\small
\caption{Hallucination detection F1 comparing \methodname against baselines, with
\mbox{Llama-4-Scout} as the KAPING answering LLM. Bold and underlined values indicate the best
and second-best result per column, respectively. Avg.\ Rank is the average rank of each method
across the three datasets (lower is better). Values with $\pm$ denote mean $\pm$ std over 3
runs. G, RS, and RT denote the retrieved subgraph, reasoning summary, and reasoning triples
provided to the LLM judge, respectively.}
\label{tab:results_summary}
\centering
\begin{tabular}{llllc}
\toprule
\textbf{Method} & \textbf{WebQSP} & \textbf{CWQ} & \textbf{PUGG} & \textbf{Avg. Rank} \\
\midrule
${\methodname\ }_{\text{(GraphTransformer)}}$ & \textbf{82.0 $\pm$ 0.7} & \underline{87.4 $\pm$ 0.2} & \underline{84.3 $\pm$ 1.0} & 1.7 \\
${\methodname\ }_{\text{(GINE)}}$ & \underline{81.2 $\pm$ 0.3} & 87.1 $\pm$ 0.3 & 83.1 $\pm$ 1.0 & 3.3 \\
${\methodname\ }_{\text{(GAT)}}$ & 79.8 $\pm$ 1.2 & 86.3 $\pm$ 0.7 & 82.7 $\pm$ 2.7 & 4.3 \\
\midrule
${\text{SelfCheckGPT\ }}_{\text{(N=10)}}$ & 61.2 & 84.0 & \textbf{85.5} & 4.3 \\
Most Frequent & 65.9 & 83.5 & 83.5 & 5.0 \\
${\text{RoBERTa\ }}_{\text{(fine-tuned)}}$ & 63.7 $\pm$ 5.2 & \textbf{87.7 $\pm$ 0.7} & 76.7 $\pm$ 2.6 & 5.0 \\
${\text{SelfCheckGPT\ }}_{\text{(N=5)}}$ & 58.9 & 82.3 & 83.9 & 6.7 \\
${\text{GPT-5-mini\ }}_{\text{(G+RS+RT)}}$ & 59.2 & 84.2 & 81.2 & 6.7 \\
${\text{GPT-5-mini\ }}_{\text{(G)}}$ & 55.9 & 82.6 & 82.6 & 8.0 \\
${\text{Llama-4-Scout\ }}_{\text{(G+RS)}}$ & 36.5 & 61.3 & 61.3 & 11.3 \\
${\text{Llama-4-Scout\ }}_{\text{(G)}}$ & 36.7 & 60.2 & 68.8 & 11.3 \\
${\text{Llama-4-Scout\ }}_{\text{(G+RS+RT)}}$ & 41.0 & 61.2 & 60.8 & 11.3 \\
Random & 49.6 $\pm$ 1.0 & 58.3 $\pm$ 0.5 & 58.8 $\pm$ 1.8 & 12.0 \\
\bottomrule
\end{tabular}
\end{table}

The other leading baselines are dataset-dependent: RoBERTa narrowly
leads on CWQ but trails \methodname by $18.3$ F1 points on WebQSP.
Despite this dataset-specific lead, RoBERTa's average rank across
the three datasets is only $5.0$, placing it fifth overall (tied
with the trivial Most Frequent baseline), which indicates that its
performance varies substantially across datasets. Similarly,
SelfCheckGPT ($N{=}10$) leads on PUGG but trails by $20.8$ points
on WebQSP and $3.4$ points on CWQ. GPT-5-mini is the strongest LLM-judge baseline on
PUGG but trails \methodname by $22.8$ points on WebQSP. In contrast,
\methodname ranks first on WebQSP and is within $0.3$--$1.2$ F1 points
of the leader on CWQ and PUGG, consistently placing among the top
methods across fine-tuned text-encoder, sampling-based, and
LLM-judge baselines. This highlights the advantage of graph-based representation.

LLM-based judges are less competitive overall. On WebQSP, all
Llama-4-Scout variants fall below the Random baseline (F1
$36.5$--$41.0$ vs.\ $49.6$), and GPT-5-mini reaches $59.2$, compared
with $82.0$ for \methodname. GPT-5-mini is stronger on CWQ and PUGG but
remains below \methodname on all three datasets. Across LLM judges, richer context does not consistently improve detection.
On CWQ and PUGG, over $71\%$ of answers are hallucinated
(Table~\ref{tab:dataset_overview}), which inflates F1 for trivial
strategies: Most Frequent reaches $83.5$ F1 on both datasets,
but its AUC-PR equals the class frequency ($71.7$ on CWQ,
$71.6$ on PUGG), compared to $94.2$ and $90.0$ for \methodname
(Tables~\ref{tab:results_cwq},~\ref{tab:results_pugg}).

WebQSP consists of naturally formulated user queries, whereas CWQ
questions are derived programmatically from logical reasoning templates
, and a portion of PUGG questions share a similar structure. We
hypothesize that judge-based methods benefit from the regular reasoning
patterns inherent to template-derived questions; naturally phrased
queries may provide weaker structural regularities. Other differences,
including class balance and retrieval coverage, may also contribute.
\methodname performs well across all three datasets regardless of this
distinction.

In the main comparison, SelfCheckGPT improves with sample count
($N{=}10$ over $N{=}5$) and outperforms all Llama-4-Scout judge
variants, but requires the most LLM calls per example.

LLM judges are strongly precision-skewed (\mbox{GPT-5-mini}: $85.3$ precision at $45.3$ recall on WebQSP),
missing many hallucinated nodes. Contrastingly, \methodname
is more balanced, and the F1 gap is driven primarily by higher recall.
Moreover, \methodname's precision--recall trade-off can be adjusted via
the classification threshold, an option unavailable to LLM judges
(Appendix~\ref{sec:additional_results-llama_full}).

Results with \mbox{Gemma 4 E4B} and \mbox{GPT OSS 20B} as the KAPING
answering LLM are reported in
Appendix~\ref{sec:additional_results-additional_llms}
(Table~\ref{tab:results_summary_additional_llms}). Consistent with
\mbox{Llama-4-Scout} results, \methodname ranks first on 5 of 6
dataset--LLM combinations; on the remaining one, the best \methodname
variant trails the best baseline by $0.5$ F1.
Results under top-$K$ retrieval are in
Appendix~\ref{sec:additional_results-topk}
(Table~\ref{tab:results_topk_summary}): all three \methodname variants
occupy the top three positions by average rank.
Results on the \mbox{G-Retriever} pipeline are reported in
Appendix~\ref{sec:additional_results-gretriever}
(Table~\ref{tab:results_gretriever_summary}): \methodname (GINE) and
\methodname (GraphTransformer) tie for the second-best average rank
($3.0$), behind only expensive \mbox{GPT-5-mini}.
Averaged over the main benchmark, top-$K$ retrieval, additional
answering LLMs, and \mbox{G-Retriever}, \methodname (GINE) has the
best average rank ($2.4$).\footnote{Each method's own rank, averaged
with equal weight over Tables~\ref{tab:results_summary},
\ref{tab:results_topk_summary},
\ref{tab:results_summary_additional_llms},
and~\ref{tab:results_gretriever_summary}.}

\subsection{Hallucination Correction}
\label{sec:results-hallucination_correction}

KBQA performance before and after hallucination-aware iterative refinement
is summarized in Table~\ref{tab:refinement_kbqa_summary}
and full per-dataset results are in Appendix~\ref{sec:additional_results-hallucination_correction}. We observe
consistent gains across all three datasets: mean F1 improves by
$+13.8$ points and Exact Match by $+17.4$ points on average.
Exact Match requires the predicted answer set to match the
gold set exactly, so any missing or spurious node counts as a failure. Its
increase shows that refinement produces fully correct answer sets for
more evaluated questions. These results demonstrate that \methodname's
node-level feedback can support targeted answer correction.

 \begin{table}[!htb]
\small
\caption{KBQA answer quality before (\underline{Initial}) and after (\underline{Refined}) iterative refinement guided by \methodname, measured by F1 and Exact Match (\underline{EM}). $\Delta$ is the score improvement.}
\label{tab:refinement_kbqa_summary}
\centering
\begin{tabular}{rrrrrrr}
\toprule
 & \multicolumn{2}{c}{\textbf{WebQSP}} & \multicolumn{2}{c}{\textbf{CWQ}} & \multicolumn{2}{c}{\textbf{PUGG}} \\
 & \textbf{F1} & \textbf{EM} & \textbf{F1} & \textbf{EM} & \textbf{F1} & \textbf{EM} \\
\midrule
\textbf{Initial} & 59.5 & 42.8 & 57.2 & 51.1 & 53.1 & 45.3 \\
\textbf{Refined} & 72.5 & 60.4 & 71.7 & 68.8 & 67.2 & 62.3 \\
\textbf{$\Delta$} & +13.0 & +17.6 & +14.5 & +17.6 & +14.0 & +16.9 \\
\bottomrule
\end{tabular}
\end{table}

\begin{figure}[!htb]
    \centering
    \includegraphics[width=0.75\textwidth]{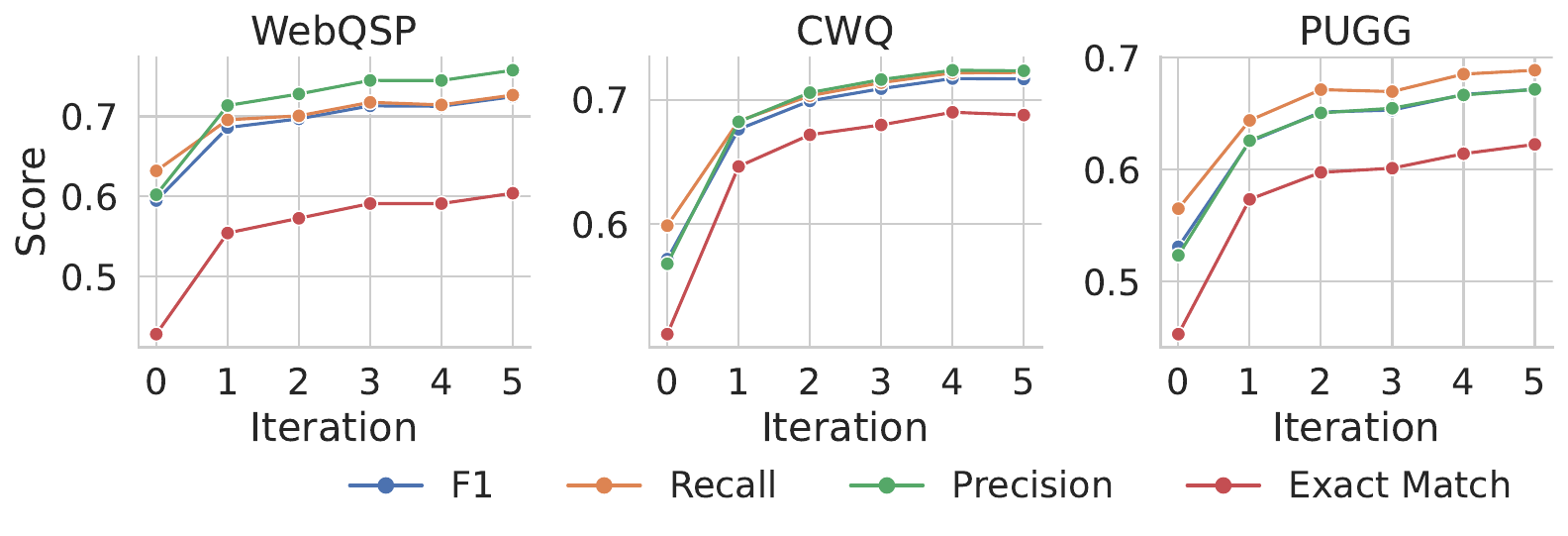}
    \caption{KBQA metrics per refinement step across three datasets. 
    Step~0 is the initial unrefined prediction; the largest gains occur at step~1, 
    after which the curve flattens.}
    \label{fig:refinement_metrics}
\end{figure}

\begin{figure}[!htb]
    \centering
    \includegraphics[width=0.75\textwidth]{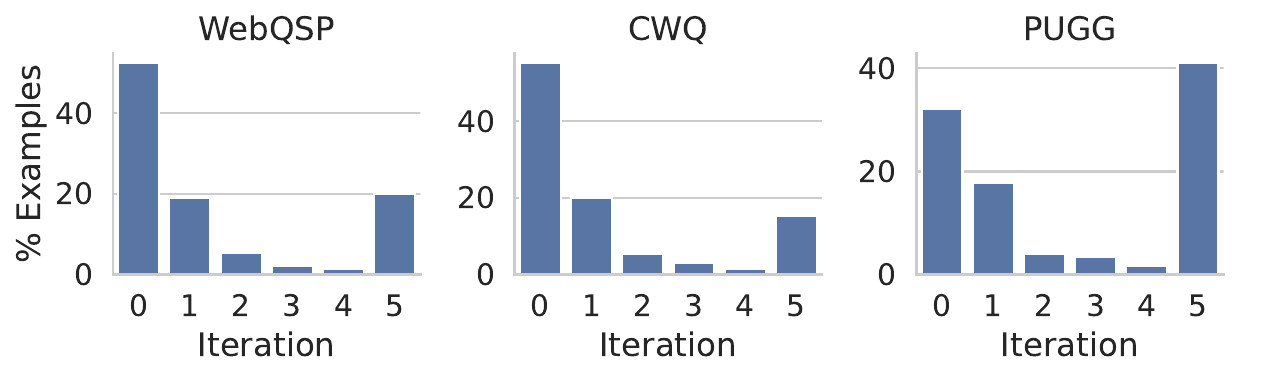}
    \caption{Distribution of refinement iterations per example across three datasets. 
    Step~0 denotes examples in which \methodname flags no hallucinated answer node before any refinement; 
    Step~5 denotes examples that reached the iteration cap.}
    \label{fig:refinement_iter_distribution}
\end{figure}

The metrics per refinement step are shown in Figure~\ref{fig:refinement_metrics}.
The largest gains occur after the first refinement step, after which the
improvement curve flattens. The distribution of iterations per example
(Figure~\ref{fig:refinement_iter_distribution}) reveals how many corrections
each example needed. Step~0 denotes examples in which \methodname flags no
hallucinated answer node at the initial step. Among examples resolved within the
iteration cap, the majority resolve at the first refinement iteration,
and the count drops quickly at later steps. A substantial fraction of
examples reach the iteration cap without converging, indicating
examples for which the detector continues to flag at least one answer.
Here, \emph{resolved} means that no answer is flagged. Among examples satisfying this stopping
criterion within the cap, most stop after one refinement iteration. Practitioners
can consider a single refinement step to reduce inference costs;
the quality trade-off is shown in Figure~\ref{fig:refinement_metrics}. 

\section{Conclusion}
\label{sec:conclusion}

We formulated hallucination detection in KBQA as answer-node
classification on retrieved KG subgraphs and proposed \methodname, a
lightweight black-box graph-based detector. Our method achieves the best
average rank ($1.7$) across all three benchmarks, ranking first on
WebQSP and within $0.3$--$1.2$ F1 points of the best on CWQ and PUGG.
The second-best method on average, SelfCheckGPT ($N{=}10$), is a
sampling-based baseline that requires $10$ LLM calls per example. In
contrast, \methodname needs only a single feed-forward pass and uses
${\sim}305\times$ fewer parameters than LLM-based baselines,
requiring zero LLM calls at detection time. The detector generalizes
across question types and languages.

Beyond detection, the node-level output of \methodname is actionable:
feeding flagged answer nodes back to the KBQA LLM as targeted correction
signals yields consistent downstream KBQA gains across all three
datasets, with most of the improvement captured in a single refinement step.
Across separately trained configurations using different answering LLMs,
retrieval strategies, and KBQA pipelines, \methodname (GINE)
achieves the best average rank ($2.4$).
Together, these results establish explicit graph-based structural verification as a
practical and efficient path toward more reliable LLM-based KBQA systems.

\section*{Limitations and Future Work}
\label{sec:limitations_and_future_work}

\methodname requires training examples with hallucination labels. 
These can be derived from any KBQA dataset by comparing
answer nodes against gold answers or by labeling hallucinations directly. 
Because we perform
node-level classification on the retrieved subgraph, it cannot handle
cases where the LLM abstains with an \emph{unknown} prediction. Such
responses are not represented as nodes in the graph. However,
defining what counts as a hallucinated abstention is non-trivial:
LLMs are generally expected to abstain when the answer is genuinely
uncertain. If a gold answer is never retrieved into the subgraph, the
refinement loop (Section~\ref{sec:hallucination_correction}) cannot
recover it either, since it only re-queries the same subgraph;
extending it with subgraph expansion or re-retrieval could address
this and is left for future work.
The answering LLM also produces a free-form reasoning summary and
supporting triples, used by the LLM-as-judge baselines. We experimented with
marking reasoning triples and incorporating the summary embedding, but
neither approach yielded consistent gains. We hypothesize that the reasoning signal is
already recoverable from the subgraph. Whether more sophisticated
fusion methods could exploit these signals remains open for future
work.

\paragraph{Broader impacts} \methodname aims to improve reliability of
LLM-based KBQA systems in high-stakes domains.
As any ML system can produce incorrect predictions, it should augment
rather than replace human factual verification.

\begin{ack}

The research was supported by the Department of Artificial Intelligence
and the Wrocław Centre for Networking and Supercomputing at Wrocław
University of Science and Technology. AS was funded by the National
Science Centre, Poland, under CHIST-ERA Open \& Re-usable Research
Data \& Software (grant number 2022/04/Y/ST6/00183). TK was supported
by the AITAX project under grant FENG.02.02-IP.05-0314/23, Action 2.2
FIRST TEAM, European Funds for a Modern Economy Programme 2021--2027
(FENG).
\end{ack}

\bibliographystyle{plainnat}
\bibliography{main}

\clearpage
\appendix

\section*{Appendix}

This appendix provides supplementary material organized as follows:

\textbf{Appendix \ref{sec:implementation_details}:} Implementation details.

\textbf{Appendix \ref{sec:additional_experiments}:} Ablation studies on graph encoder
components, input modalities, virtual question-node connectivity, and graph encoder
architecture, the effect of training with KBQA-derived labels, and classifier
calibration.

\textbf{Appendix \ref{sec:model_size_and_computation_demands}:} Model size comparison
and computation demands across methods.

\textbf{Appendix \ref{sec:additional_results}:} Hallucination detection results with
additional answering LLMs (Gemma 4 E4B, GPT OSS 20B), top-$K$ retrieval,
the \mbox{G-Retriever} pipeline, full detection metrics, and
KBQA correction metrics including Exact Match.

\textbf{Appendix \ref{sec:additional_statistics}:} Dataset and subgraph statistics,
including subgraph structure, answer count distributions, and LLM abstention and hallucination
patterns.

\section{Implementation Details}
\label{sec:implementation_details}

\paragraph{LLM settings} For KAPING, the answering LLM is one of
\mbox{Llama-4-Scout} \citep{meta2025llama4},
\mbox{Gemma 4 E4B} \citep{google2025gemma4}, or
\mbox{GPT OSS 20B} \citep{openai2025gptoss}, each at temperature
$0.1$ to favor the most probable outputs while allowing retries on
output-parsing errors. \mbox{Llama-4-Scout} is the primary
answering LLM in all main-paper experiments; \mbox{Gemma 4 E4B} and
\mbox{GPT OSS 20B} are used as additional answering LLMs in
Appendix~\ref{sec:additional_results-additional_llms}.
\mbox{Llama-4-Scout} is also used as the LLM-as-judge baseline and
for answer refinement. SelfCheckGPT draws its $N$ samples from the
answering LLM of that setting, at temperature $1.0$. \mbox{GPT-5-mini} is
queried with temperature $1$ and default reasoning effort
(\texttt{medium}); temperature $1$ is the only value permitted by the
API when reasoning is enabled.

\paragraph{\methodname training} Node, edge, and question embeddings
are produced by a frozen Sentence Transformer \citep{reimers2019sbert}:
\texttt{all-roberta-large-v1} \citep{reimers2019sbert}
for WebQSP and CWQ, and multilingual
\texttt{mmlw-retrieval-roberta-large} \citep{dadas2024pirb} for PUGG
(in both cases $d = 1024$). The same encoders are used for top-$K$
embedding retrieval. Node features are
obtained by encoding the node label, and edge features by encoding
the relation label. Dimensions of marking embeddings for topic entities and answer nodes
are set to $d_T = d_A = 20$, plus a 1-dimensional virtual-question-node
indicator, giving node features of size $1065$. The learnable
embeddings for edges between the virtual question node and the topic
entities are sized to $1024$.
The graph encoder is instantiated as a GraphTransformer
\citep{dwivedi2021generalization,shi2021maskedlabel}, a GAT \citep{velivckovic2018gat}, or GINE
\citep{hu2020strategies}. All variants use $2$ layers and
$256$-dimensional hidden representations; GraphTransformer and GAT
additionally use $8$ attention heads.
We train with Adam ($\text{lr}{=}10^{-3}$, weight decay $10^{-4}$), batch size
$32$, for up to $300$ epochs with early stopping (patience $50$),
minimizing cross-entropy loss over the factual/hallucinated logits.
Checkpoint selection and early stopping both monitor the validation F1 score.
All hyperparameter values were chosen empirically.
The implementation uses
PyTorch Geometric \citep{fey2019fast} and PyTorch Lightning \citep{falcon2019pytorchlightning}.

\paragraph{Compute resources} LLM inference (\mbox{Llama-4-Scout},
\mbox{Gemma 4 E4B}, and \mbox{GPT OSS 20B}) was performed using vLLM on
4~$\times$ H100 GPUs. \methodname training was conducted on an NVIDIA A40
GPU (see Appendix~\ref{sec:model_size_and_computation_demands} for details).

\section{Additional Experiments}
\label{sec:additional_experiments}

\subsection{Effect of Graph Encoder Components}
\label{sec:additional_experiments-graph_encoder_components}

\begin{table}[!htb]
\small
\caption{Ablation of graph encoder components on hallucination detection F1.
$\Delta$ is the F1 change in percentage points relative to the full
\methodname model; negative values indicate performance loss from
removing the component. Bold and underlined values indicate the best
and second-best F1 per dataset, respectively. Mean $\pm$ std over 3 runs.}
\label{tab:ablation_gnn}
\centering
\begin{tabular}{llrlrlr|r}
\toprule
\textbf{Variant} & \textbf{WebQSP} & \textbf{$\Delta$} & \textbf{CWQ} & \textbf{$\Delta$} & \textbf{PUGG} & \textbf{$\Delta$} & \textbf{Avg. $\Delta$} \\
\midrule
full & \textbf{82.0 $\pm$ 0.7} &  & \underline{87.4 $\pm$ 0.2} &  & \textbf{84.3 $\pm$ 1.0} &  & +0.0 \\
w/o marking & 80.1 $\pm$ 0.9 & -1.9 & \textbf{87.9 $\pm$ 0.3} & +0.5 & 83.4 $\pm$ 1.3 & -0.9 & -0.8 \\
w/o virtual q. node & \underline{81.6 $\pm$ 0.2} & -0.5 & 86.6 $\pm$ 1.0 & -0.8 & \underline{83.9 $\pm$ 0.4} & -0.4 & -0.5 \\
w/o marking \& virtual q. node & 77.8 $\pm$ 0.9 & -4.2 & 86.5 $\pm$ 0.5 & -0.9 & 82.7 $\pm$ 0.8 & -1.6 & -2.2 \\
\bottomrule
\end{tabular}
\end{table}

To justify the proposed graph-encoder design, we ablate node marking and
the virtual question node independently and in combination.
Table~\ref{tab:ablation_gnn} presents the results. 
Removing node marking (topic entity and answer node trainable embeddings)
reduces F1 by $0.8$ points on average, while removing the virtual question
node reduces it by $0.5$ points; their combined removal reduces F1 by
$2.2$ points, exceeding
the sum of the individual effects, indicating the two are
mutually reinforcing. CWQ is notably more robust across variants, whereas WebQSP
shows the largest sensitivity to ablation.

\subsection{Effect of Input Modalities}
\label{sec:additional_experiments-input_modalities}

\begin{table}[!htb]
\small
\caption{Ablation of input modalities on hallucination detection F1.
$\Delta$ is the F1 change in percentage points relative to the full
\methodname model (graph + question text); negative values indicate
performance loss from removing the modality. The reference row is shown
in bold. RS is the reasoning summary produced by the answering LLM and
was not used in \methodname. Mean $\pm$ std over 3 runs.}
\label{tab:ablation_modality}
\centering
\begin{tabular}{llrlrlr|r}
\toprule
\textbf{Variant} & \textbf{WebQSP} & \textbf{$\Delta$} & \textbf{CWQ} & \textbf{$\Delta$} & \textbf{PUGG} & \textbf{$\Delta$} & \textbf{Avg. $\Delta$} \\
\midrule
full (graph + text) & \textbf{82.1 $\pm$ 0.4} &  & \textbf{87.6 $\pm$ 0.5} &  & \textbf{83.7 $\pm$ 0.6} &  & +0.0 \\
graph only & 80.6 $\pm$ 0.6 & -1.5 & 86.3 $\pm$ 0.1 & -1.3 & 83.1 $\pm$ 0.6 & -0.6 & -1.1 \\
question only & 70.8 $\pm$ 2.3 & -11.4 & 83.0 $\pm$ 0.5 & -4.6 & 81.7 $\pm$ 0.3 & -2.0 & -6.0 \\
question+RS & 70.6 $\pm$ 2.0 & -11.5 & 83.3 $\pm$ 0.6 & -4.3 & 81.2 $\pm$ 0.3 & -2.5 & -6.1 \\
\bottomrule
\end{tabular}
\end{table}

We investigate how much each input modality contributes to detection and
whether the LLM's reasoning summary (RS) (not used by \methodname) can
substitute for graph structure. Table~\ref{tab:ablation_modality} reports
the results. Ablating text
features (graph only) reduces F1 by $1.1$ points on average;
removing the graph encoder (question only) reduces F1 by $6.0$ points.
These ablations support the contribution of the graph component. Adding the
reasoning summary to the text input (question+RS) yields no improvement over
question text alone ($-6.1$ vs.\ $-6.0$ points). Thus, adding the
reasoning summary does not improve this text-only variant in our
experiments. WebQSP is the most sensitive dataset for this ablation.

\subsection{Virtual Question-Node Connectivity}
\label{sec:additional_experiments-virtual_node_connectivity}

\begin{table}[!htb]
\small
\caption{Virtual question-node connectivity ablation on hallucination
detection F1. Bold and underlined values indicate the best and
second-best result per column, respectively. Mean $\pm$ std over 3 runs.}
\label{tab:ablation_virtual_node}
\centering
\begin{tabular}{llll}
\toprule
\textbf{Variant} & \textbf{WebQSP} & \textbf{CWQ} & \textbf{PUGG} \\
\midrule
Topic entities & \underline{82.0 $\pm$ 0.7} & 87.4 $\pm$ 0.2 & 84.3 $\pm$ 1.0 \\
Answer nodes & \textbf{82.1 $\pm$ 0.6} & \underline{87.5 $\pm$ 0.2} & \underline{84.4 $\pm$ 0.6} \\
All nodes & 81.4 $\pm$ 1.3 & \textbf{88.4 $\pm$ 0.2} & \textbf{84.8 $\pm$ 0.5} \\
\bottomrule
\end{tabular}
\end{table}

To inject question semantics, \methodname connects the virtual
question node to topic entities only. We compare this design against
connecting the virtual question node to all retrieved nodes or to
the proposed answer nodes, using the GraphTransformer encoder.
Table~\ref{tab:ablation_virtual_node} presents the results. The
answer-node and topic-entity variants have similar F1 scores.
Connecting to all nodes improves CWQ by $1.0$ F1 point and PUGG by
$0.5$ points but reduces WebQSP by $0.6$ points. Connecting to all
retrieved nodes also adds $2|V|$ directed edges instead of $2|T|$,
raising compute and memory cost on heavy-tailed graphs such as
PUGG's. Topic-entity and answer-node
connections yield similar F1 scores; we use topic entities throughout
\methodname.

\subsection{Graph Encoder Architecture}
\label{sec:additional_experiments-gnn_architecture}

\begin{table}[!htb]
\small
\caption{Graph encoder architecture ablation on hallucination
detection F1. Bold and underlined values indicate the best and
second-best result per column, respectively. Avg.\ Rank is the
average rank of each variant across the three datasets (lower is
better). Mean $\pm$ std over 3 runs.}
\label{tab:ablation_gnn_architecture}
\centering
\begin{tabular}{llllc}
\toprule
\textbf{Variant} & \textbf{WebQSP} & \textbf{CWQ} & \textbf{PUGG} & \textbf{Avg. Rank} \\
\midrule
GraphTransformer & \textbf{82.0 $\pm$ 0.7} & \textbf{87.4 $\pm$ 0.2} & \textbf{84.3 $\pm$ 1.0} & 1.0 \\
GINE & \underline{81.2 $\pm$ 0.3} & \underline{87.1 $\pm$ 0.3} & 83.1 $\pm$ 1.0 & 2.7 \\
GraphSAGE & 79.2 $\pm$ 0.7 & 86.5 $\pm$ 0.2 & 83.6 $\pm$ 0.5 & 3.3 \\
GAT & 79.8 $\pm$ 1.2 & 86.3 $\pm$ 0.7 & 82.7 $\pm$ 2.7 & 4.0 \\
GCN & 75.2 $\pm$ 1.3 & 84.9 $\pm$ 0.7 & \underline{83.7 $\pm$ 0.7} & 4.0 \\
\bottomrule
\end{tabular}
\end{table}

We compare the GraphTransformer encoder used in \methodname against
four alternatives that vary along two axes: GAT, like GraphTransformer,
aggregates via attention, while GINE, GraphSAGE, and GCN aggregate
without attention; and GAT and GINE, like GraphTransformer, condition
message passing on edge features, while GraphSAGE and GCN do not.
Table~\ref{tab:ablation_gnn_architecture} presents the results.
GraphTransformer achieves the best F1 on all three datasets, followed
on average by GINE, GraphSAGE, and GAT, with GCN weakest overall.
Notably, GCN is the second-best variant on PUGG despite lacking edge
features, and GraphSAGE and GCN together degrade only moderately
rather than collapsing. This suggests that node topology and node-level
marking already carry most of the discriminative signal. However, these
architectures also differ in other ways beyond their use of edge features,
so this comparison does not isolate edge features' individual contribution.
GraphTransformer's consistent lead across datasets motivates its use
as the main graph encoder in \methodname.

\subsection{Training with KBQA-Derived Labels}
\label{sec:additional_experiments-training_with_kbqa_dataset}

A natural alternative to our approach is to train the same model
to predict correct answer nodes using KBQA gold labels, then flag
LLM-proposed answers as hallucinated whenever the model judges them to be incorrect.
Since the KBQA datasets are considerably larger than our
hallucination detection dataset, we subsample each KBQA split to match its size
and class balance. We repeat the procedure with three random seeds for each training run
(Table~\ref{tab:kbqa_dataset_stats}).

\begin{table}[!htb]
\small
\caption{Hallucination detection when training \methodname on
hallucination labels (Hall.\ Det.) versus KBQA-derived labels (KBQA).
Bold marks the better value in each metric. Mean $\pm$ std over 3 runs.}
\label{tab:kbqa_results}
\centering
\begin{tabular}{lllllll}
\toprule
\textbf{Dataset} & \textbf{Train. Data} & \textbf{F1} & \textbf{Precision} & \textbf{Recall} & \textbf{Accuracy} & \textbf{AUC-PR} \\
\midrule
\multirow{2}{*}{\textbf{WebQSP}} & Hall. Det. & \textbf{82.0 $\pm$ 0.7} & \textbf{84.9 $\pm$ 2.2} & 81.3 $\pm$ 2.9 & \textbf{85.9 $\pm$ 0.1} & \textbf{91.4 $\pm$ 0.1} \\
 & KBQA & 81.4 $\pm$ 0.9 & 81.6 $\pm$ 1.9 & \textbf{83.2 $\pm$ 1.0} & 85.1 $\pm$ 0.8 & 90.6 $\pm$ 0.2 \\
 \midrule
\multirow{2}{*}{\textbf{CWQ}} & Hall. Det. & \textbf{87.4 $\pm$ 0.2} & 85.8 $\pm$ 0.2 & \textbf{90.2 $\pm$ 0.5} & \textbf{83.5 $\pm$ 0.2} & \textbf{94.2 $\pm$ 0.3} \\
 & KBQA & 81.7 $\pm$ 0.7 & \textbf{88.1 $\pm$ 0.2} & 77.7 $\pm$ 1.0 & 78.8 $\pm$ 0.5 & 92.5 $\pm$ 0.1 \\
\midrule
\multirow{2}{*}{\textbf{PUGG}} & Hall. Det. & \textbf{84.3 $\pm$ 1.0} & 78.1 $\pm$ 3.4 & \textbf{93.5 $\pm$ 4.8} & \textbf{78.8 $\pm$ 1.9} & \textbf{90.0 $\pm$ 0.6} \\
 & KBQA & 71.3 $\pm$ 5.7 & \textbf{84.1 $\pm$ 2.0} & 65.3 $\pm$ 9.3 & 69.8 $\pm$ 3.7 & 85.4 $\pm$ 0.2 \\
\bottomrule
\end{tabular}
\end{table}

\begin{table}[!htb]
\small
\caption{Training set statistics for hallucination-specific (\underline{Hall.\ Det.}) and KBQA-derived supervision (\underline{KBQA-original}: original split; \underline{KBQA-sampled}: subsampled to match Hall.\ Det.\ size for fair comparison). \underline{\#Instances}: total labeled nodes; \underline{\#Pos.}/\underline{\#Neg.}: hallucinated/correct instances; \underline{\%Pos.}: class imbalance; \underline{\#Graphs}: unique subgraphs. Mean $\pm$ std over sampling runs.}
\label{tab:kbqa_dataset_stats}
\centering
\begin{tabular}{llrrrr r@{}l}
\toprule
\textbf{Dataset} & \textbf{Variant} & \textbf{\#Instances} & \textbf{\#Pos.} & \textbf{\#Neg.} & \textbf{\%Pos.} & \multicolumn{2}{c}{\textbf{\#Graphs}} \\
\midrule
\multirow{3}{*}{\textbf{WebQSP}} & Hall. Det. & 8677 & 4592 & 4085 & 52.9 & 2332 & \\
 & KBQA-original & 43768 & 38115 & 5653 & 87.1 & 2826 & \\
 & KBQA-sampled & 8677 & 4592 & 4085 & 52.9 & 2387 & $\pm8.7$ \\
\midrule
\multirow{3}{*}{\textbf{CWQ}} & Hall. Det. & 33206 & 20833 & 12373 & 62.7 & 17223 & \\
 & KBQA-original & 456141 & 438020 & 18121 & 96.0 & 27631 & \\
 & KBQA-sampled & 33206 & 20833 & 12373 & 62.7 & 16154 & $\pm46.9$ \\
\midrule
\multirow{3}{*}{\textbf{PUGG}} & Hall. Det. & 5467 & 3916 & 1551 & 71.6 & 2777 & \\
 & KBQA-original & 350208 & 348000 & 2208 & 99.4 & 3589 & \\
 & KBQA-sampled & 5467 & 3916 & 1551 & 71.6 & 1683 & $\pm5.9$ \\
\bottomrule
\end{tabular}
\end{table}

Table~\ref{tab:kbqa_results} shows our method outperforms the
KBQA-supervised model on all three datasets despite identical training
set sizes, indicating that hallucination-specific labels provide a
stronger training signal than KBQA gold labels for this task.

Annotating proposed answers may reduce the scope of annotation:
an annotator judges a candidate answer rather than identifying every
correct answer in the KG. However, this can still require domain
expertise and access to evidence beyond the retrieved subgraph.

\subsection{Calibration}
\label{sec:additional_experiments-calibration}

\begin{figure*}[!htb]
    \centering
    \includegraphics[width=0.7\textwidth]{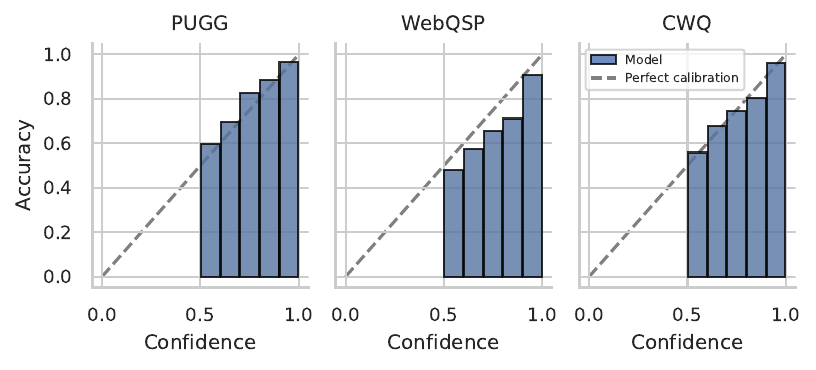}
    \caption{Reliability diagrams of \methodname on the test splits
    ($10$ bins, $3$ seeds pooled). The dashed line denotes perfect
    calibration.}
    \label{fig:calibration_reliability}
\end{figure*}

Expected Calibration Error (ECE) is the average gap between the
classifier's confidence and the fraction of correct predictions. With
\mbox{Llama-4-Scout} as the answering LLM, pooling the $3$ seeds,
test ECE of \methodname on confidence in $10$ bins is $0.045$
on PUGG and $0.023$ on CWQ, despite over $71\%$ hallucinated answers
among non-abstaining nodes (Table~\ref{tab:dataset_overview}), and
$0.084$ on the more balanced WebQSP.
Figure~\ref{fig:calibration_reliability} shows bin accuracy tracking
the diagonal on PUGG and CWQ, with mild overconfidence on WebQSP.
No post-hoc calibration is applied. The decision threshold can be
tuned on the validation split to match the precision--recall trade-off
required in deployment.

\section{Model Size and Computation Demands}
\label{sec:model_size_and_computation_demands}

Table~\ref{tab:model_size} compares model sizes and query-time LLM call
counts across methods.
\methodname (GraphTransformer) adds only $2.52$M trainable
parameters on top of a frozen sentence encoder (${\sim}358$M total
for WebQSP/CWQ and ${\sim}437$M for PUGG).
The fine-tuned RoBERTa baseline
is comparable in total size (${\sim}355$M), but all $355$M parameters
are updated during training, requiring full back-propagation through the
encoder for each target dataset. \methodname keeps this encoder frozen
and trains only the graph-encoder and MLP parameters. WebQSP and CWQ reuse the
same frozen English encoder; PUGG uses a separate frozen multilingual encoder.
LLM-based baselines are far larger: \mbox{Llama-4-Scout}
operates at $109$B parameters---$305\times$ larger than our full
inference stack using the English-language encoder (${\sim}249\times$
using Polish-language encoder); at \texttt{fp16}
precision this corresponds to ${\sim}0.7$\,GB
for \methodname with the English encoder (${\sim}0.87$\,GB with the
Polish-language encoder) versus ${\sim}218$\,GB for \mbox{Llama-4-Scout},
exceeding the capacity of a single high-end GPU.
Each LLM call requires autoregressive generation: while the final answer
spans only several tokens, \emph{thinking} enabled in \mbox{GPT-5-mini}
can consume hundreds of additional tokens per query.
SelfCheckGPT further multiplies this cost by querying the
answering LLM $N{=}5$ or $N{=}10$ times per query.
\methodname makes zero LLM calls at detection time: classification is a
single feed-forward pass through the graph encoder and a small MLP.

In practical deployment over a fixed knowledge graph, node and relation
embeddings can be precomputed and stored offline. During inference, only the
question requires encoding, after which one graph-encoder forward pass
and a batched MLP evaluation produce hallucination scores for the
proposed answer nodes.
This lightweight design makes \methodname deployable on commodity
hardware, with no multi-GPU infrastructure or external API access required.

\begin{table}[!htb]
\small
\caption{Model size and per-query \underline{LLM calls} at detection time. \underline{Encoder}: frozen SentenceTransformer. \underline{Trainable}: task-specific trainable parameters (\methodname: graph encoder + MLP; RoBERTa: full model). \underline{Total}: Encoder + Trainable. \underline{$\times$\,ours}: total parameter ratio relative to \methodname. \mbox{Llama-4-Scout} has 109B total, but 17B active parameters (MoE); SelfCheckGPT also uses \mbox{Llama-4-Scout} but in a sampling-based paradigm ($N$ calls per query). \mbox{GPT-5-mini} parameter count is undisclosed.}
\label{tab:model_size}
\centering
\begin{tabular}{llrll l}
\toprule
\textbf{Method} & \textbf{Encoder} & \textbf{LLM calls} & \textbf{Trainable} & \textbf{Total} & \textbf{$\times$ ours} \\
\midrule
${\methodname\ }_{\text{(GraphTransformer)}}$ & all-roberta-large-v1 & 0 & 2.52M & 358M & — \\
${\methodname\ }_{\text{(GraphTransformer)}}$ & mmlw-retrieval-roberta-large & 0 & 2.52M & 437M & — \\
${\methodname\ }_{\text{(GINE)}}$ & all-roberta-large-v1 & 0 & 2.47M & 358M & — \\
${\methodname\ }_{\text{(GINE)}}$ & mmlw-retrieval-roberta-large & 0 & 2.47M & 437M & — \\
${\methodname\ }_{\text{(GAT)}}$ & all-roberta-large-v1 & 0 & 1.51M & 357M & — \\
${\methodname\ }_{\text{(GAT)}}$ & mmlw-retrieval-roberta-large & 0 & 1.51M & 436M & — \\
${\text{RoBERTa\ }}_{\text{(fine-tuned)}}$ & — & 0 & 355M & 355M & ${\sim}1\times$ \\
GPT-5-mini & — & 1 & — & n/a & n/a \\
${\text{SelfCheckGPT\ }}_{\text{(N=10)}}$ & — & 10 & — & 109B & $305\times$ \\
${\text{SelfCheckGPT\ }}_{\text{(N=5)}}$ & — & 5 & — & 109B & $305\times$ \\
Llama-4-Scout & — & 1 & — & 109B & $305\times$ \\
\bottomrule
\end{tabular}
\end{table}

\section{Additional Results}
\label{sec:additional_results}

\subsection{Hallucination Detection}
\label{sec:additional_results-hallucination_detection}

\subsubsection{Additional Answering LLMs}
\label{sec:additional_results-additional_llms}

\begin{table}[!htb]
\small
\caption{Hallucination detection F1 with \mbox{Gemma 4 E4B} and
\mbox{GPT OSS 20B} as the KAPING answering LLM, sorted by average rank.
Bold and underlined values indicate the best and second-best result per
column, respectively. Avg.\ Rank is the average rank across all six
dataset--LLM combinations (lower is better). Values with $\pm$ denote
mean $\pm$ std over 3 runs. G, RS, and RT denote the retrieved subgraph,
reasoning summary, and reasoning triples provided to the LLM judge.}
\label{tab:results_summary_additional_llms}
\centering
\resizebox{\linewidth}{!}{\begin{tabular}{lllllllr}
\toprule
\multirow{2}{*}{\textbf{Method}} & \multicolumn{2}{c}{\textbf{WebQSP}} & \multicolumn{2}{c}{\textbf{CWQ}} & \multicolumn{2}{c}{\textbf{PUGG}} & \multirow{2}{*}{\shortstack[r]{\textbf{Avg.}\\\textbf{Rank}}} \\
\cmidrule(lr){2-3}\cmidrule(lr){4-5}\cmidrule(lr){6-7}
 & \textbf{Gemma 4 E4B} & \textbf{GPT OSS 20B} & \textbf{Gemma 4 E4B} & \textbf{GPT OSS 20B} & \textbf{Gemma 4 E4B} & \textbf{GPT OSS 20B} & \\
\midrule
${\methodname\ }_{\text{(GINE)}}$ & \textbf{74.6 $\pm$ 1.6} & \textbf{75.8 $\pm$ 0.6} & \textbf{77.8 $\pm$ 1.0} & \textbf{79.2 $\pm$ 1.3} & 73.8 $\pm$ 0.4 & \textbf{81.0 $\pm$ 0.6} & 1.5 \\
${\methodname\ }_{\text{(GraphTransformer)}}$ & \underline{73.1 $\pm$ 0.9} & 73.9 $\pm$ 0.4 & \underline{77.1 $\pm$ 0.1} & \underline{79.1 $\pm$ 0.3} & 74.2 $\pm$ 0.5 & 78.7 $\pm$ 0.5 & 2.8 \\
${\methodname\ }_{\text{(GAT)}}$ & \underline{73.1 $\pm$ 1.1} & \underline{74.1 $\pm$ 0.7} & 76.9 $\pm$ 0.9 & 77.3 $\pm$ 0.5 & 73.6 $\pm$ 1.2 & 78.9 $\pm$ 0.3 & 3.2 \\
\midrule
${\text{SelfCheckGPT\ }}_{\text{(N=10)}}$ & 34.4 & 63.9 & 67.5 & 74.6 & \textbf{74.7} & \underline{79.3} & 4.7 \\
${\text{SelfCheckGPT\ }}_{\text{(N=5)}}$ & 58.4 & 65.3 & 61.8 & 74.7 & 71.9 & 77.3 & 5.3 \\
${\text{RoBERTa\ }}_{\text{(fine-tuned)}}$ & 58.1 $\pm$ 1.8 & 59.4 $\pm$ 0.7 & 76.9 $\pm$ 0.8 & 78.3 $\pm$ 0.4 & 70.2 $\pm$ 2.1 & 76.0 $\pm$ 0.2 & 5.7 \\
${\text{GPT-5-mini\ }}_{\text{(G+RS+RT)}}$ & 57.9 & 53.2 & 61.9 & 65.3 & 73.3 & 76.5 & 6.7 \\
${\text{GPT-5-mini\ }}_{\text{(G)}}$ & 45.9 & 45.5 & 57.3 & 66.3 & 59.0 & 73.9 & 8.7 \\
Most Frequent & \phantom{0}0.0 & \phantom{0}0.0 & \phantom{0}0.0 & 71.0 & \underline{74.6} & 76.2 & 9.2 \\
${\text{Llama-4-Scout\ }}_{\text{(G)}}$ & 31.1 & 48.9 & 39.5 & 56.7 & 29.7 & 59.6 & 9.8 \\
Random & 43.5 $\pm$ 1.2 & 47.3 $\pm$ 0.5 & 53.0 $\pm$ 1.2 & 51.3 $\pm$ 0.6 & 53.3 $\pm$ 1.3 & 54.9 $\pm$ 0.2 & 10.0 \\
${\text{Llama-4-Scout\ }}_{\text{(G+RS+RT)}}$ & 28.4 & 45.9 & 21.4 & 47.2 & 23.5 & 59.2 & 11.3 \\
${\text{Llama-4-Scout\ }}_{\text{(G+RS)}}$ & 23.9 & 42.1 & 21.1 & 48.0 & 22.7 & 58.6 & 12.2 \\
\bottomrule
\end{tabular}}
\end{table}

Table~\ref{tab:results_summary_additional_llms} reports hallucination
detection results when KAPING is driven by \mbox{Gemma 4 E4B} or
\mbox{GPT OSS 20B} instead of \mbox{Llama-4-Scout}. Consistent with
\mbox{Llama-4-Scout} results, \methodname ranks first on 5 of 6
dataset--LLM combinations. On the remaining one (PUGG /
\mbox{Gemma 4 E4B}), it trails
SelfCheckGPT ($N{=}10$) by $0.5$ F1. All three \methodname
variants occupy the top three positions by average rank, demonstrating
robustness to the choice of answering LLM. Different answering LLMs
produce qualitatively different error profiles, including shifts in
class balance across datasets. These shifts change the difficulty
profile for trivial predictors yet leave \methodname's relative
advantage intact.

\subsubsection{Top-K Retrieval}
\label{sec:additional_results-topk}

\begin{table}[!htb]
\small
\caption{Hallucination detection F1 comparing \methodname
against baselines under top-$K{=}100$ embedding retrieval with
\mbox{KAPING} and \mbox{Llama-4-Scout}. Bold
and underlined values indicate the best and second-best result
per column, respectively. Avg.\ Rank is the average rank of each
method across the three datasets (lower is better). Values with
$\pm$ denote mean $\pm$ std over 3 runs. G, RS, and RT denote the
retrieved subgraph, reasoning summary, and reasoning triples provided
to the LLM judge, respectively.}
\label{tab:results_topk_summary}
\centering
\begin{tabular}{llllc}
\toprule
\textbf{Method} & \textbf{WebQSP} & \textbf{CWQ} & \textbf{PUGG} & \textbf{Avg. Rank} \\
\midrule
${\methodname\ }_{\text{(GINE)}}$ & \underline{82.0 $\pm$ 0.5} & \textbf{89.4 $\pm$ 0.1} & \underline{85.8 $\pm$ 1.2} & 1.7 \\
${\methodname\ }_{\text{(GAT)}}$ & 81.9 $\pm$ 0.2 & \underline{88.9 $\pm$ 0.4} & 85.4 $\pm$ 0.1 & 2.7 \\
${\methodname\ }_{\text{(GraphTransformer)}}$ & \textbf{82.6 $\pm$ 0.9} & 88.4 $\pm$ 0.5 & 84.7 $\pm$ 1.0 & 3.3 \\
\midrule
${\text{SelfCheckGPT\ }}_{\text{(N=10)}}$ & 63.5 & 87.0 & \textbf{86.8} & 3.7 \\
${\text{SelfCheckGPT\ }}_{\text{(N=5)}}$ & 61.1 & 86.1 & 85.3 & 5.7 \\
Most Frequent & 68.7 & 85.5 & 85.1 & 5.7 \\
${\text{RoBERTa\ }}_{\text{(fine-tuned)}}$ & 65.7 $\pm$ 0.3 & 86.1 $\pm$ 2.4 & 82.4 $\pm$ 0.1 & 6.3 \\
${\text{GPT-5-mini\ }}_{\text{(G+RS+RT)}}$ & 54.3 & 86.5 & 75.9 & 7.0 \\
${\text{GPT-5-mini\ }}_{\text{(G)}}$ & 48.9 & 83.6 & 67.2 & 9.3 \\
Random & 50.9 $\pm$ 0.3 & 59.8 $\pm$ 0.3 & 60.0 $\pm$ 0.9 & 9.7 \\
${\text{Llama-4-Scout\ }}_{\text{(G+RS+RT)}}$ & 37.0 & 49.7 & 50.1 & 11.0 \\
${\text{Llama-4-Scout\ }}_{\text{(G)}}$ & 28.9 & 42.7 & 42.5 & 12.0 \\
\bottomrule
\end{tabular}
\end{table}

The top-$K$ retriever keeps the $K{=}100$ triples whose embeddings are
most similar to the question embedding, selecting each triple
independently rather than as a jointly optimized connected subtree as
in PCST. This changes the connectivity, sparsity, and noise level of
the retrieved subgraphs. We regenerate the subgraphs, rerun KAPING with
\mbox{Llama-4-Scout} as the answering LLM, and retrain and evaluate all methods on the
resulting labels.
Table~\ref{tab:results_topk_summary} reports hallucination detection
results under top-$K{=}100$ embedding retrieval (F1 only). All three
\methodname variants occupy the top three positions by average rank,
consistent with PCST results, showing that \methodname's advantage is
not retrieval-specific. SelfCheckGPT ($N{=}10$) is the best-performing
baseline (average rank $3.7$), even outperforming all \methodname
variants on PUGG, but it still trails \methodname (GINE) by $18.5$ F1
points on WebQSP and requires $10$ answering-LLM calls per question
at detection time. Notably, LLM-based judges degrade more
substantially under top-$K$ retrieval than \methodname: all
\mbox{Llama-4-Scout} and \mbox{GPT-5-mini} variants fall below
\methodname by clear margins, and \mbox{Llama-4-Scout} (G) trails even the Random baseline. 
This suggests that graph
topology provides a more stable verification signal than prompt-level
context when the retrieved subgraph is a flat top-$K$ neighborhood
rather than a PCST-selected compact tree.

\subsubsection{G-Retriever Pipeline}
\label{sec:additional_results-gretriever}

\begin{table}[!htb]
\small
\caption{Hallucination detection F1 and AUC-PR comparing
\methodname against baselines under the \mbox{G-Retriever} KBQA
pipeline. Bold and underlined values indicate the best and
second-best result per column, respectively. Avg.\ Rank is the
average rank of each method across the three datasets, calculated
over F1 (lower is better). Values with $\pm$ denote mean $\pm$ std over 3 runs.
G denotes the retrieved subgraph provided to the LLM judge.}
\label{tab:results_gretriever_summary}
\centering
\resizebox{\linewidth}{!}{%
\begin{tabular}{lllllllr}
\toprule
\multirow{2}{*}{\textbf{Method}} & \multicolumn{2}{c}{\textbf{WebQSP}} & \multicolumn{2}{c}{\textbf{CWQ}} & \multicolumn{2}{c}{\textbf{PUGG}} & \multirow{2}{*}{\shortstack[r]{\textbf{Avg.}\\\textbf{Rank}}} \\
\cmidrule(lr){2-3}\cmidrule(lr){4-5}\cmidrule(lr){6-7}
 & \textbf{F1} & \textbf{AUC-PR} & \textbf{F1} & \textbf{AUC-PR} & \textbf{F1} & \textbf{AUC-PR} & \\
\midrule
${\text{GPT-5-mini\ }}_{\text{(G)}}$ & \textbf{85.9} & 82.9 & \textbf{88.4} & 86.9 & \textbf{90.5} & 92.3 & 1.0 \\
${\methodname\ }_{\text{(GINE)}}$ & 77.3 $\pm$ 0.1 & \underline{83.0 $\pm$ 0.8} & 85.9 $\pm$ 0.7 & 91.2 $\pm$ 0.9 & \underline{87.3 $\pm$ 0.2} & \underline{90.7 $\pm$ 0.8} & 3.0 \\
${\methodname\ }_{\text{(GraphTransformer)}}$ & \underline{77.4 $\pm$ 0.6} & \textbf{85.7 $\pm$ 0.7} & 86.4 $\pm$ 0.4 & \underline{93.1 $\pm$ 0.4} & 86.9 $\pm$ 0.4 & 90.4 $\pm$ 1.0 & 3.0 \\
${\text{RoBERTa\ }}_{\text{(fine-tuned)}}$ & 71.1 $\pm$ 4.6 & 78.3 $\pm$ 13.3 & \underline{86.6 $\pm$ 0.7} & \textbf{95.2 $\pm$ 0.4} & 86.9 $\pm$ 0.3 & \textbf{93.5 $\pm$ 0.5} & 3.7 \\
${\methodname\ }_{\text{(GAT)}}$ & 76.2 $\pm$ 0.4 & \underline{83.0 $\pm$ 0.3} & 85.3 $\pm$ 0.3 & 91.1 $\pm$ 0.0 & 86.5 $\pm$ 0.0 & 88.4 $\pm$ 1.4 & 4.7 \\
${\text{Llama-4-Scout\ }}_{\text{(G)}}$ & 73.4 & 70.2 & 71.4 & 81.2 & 69.0 & 83.0 & 6.3 \\
Most Frequent & 68.6 & 52.2 & 82.4 & 70.1 & 85.5 & 74.7 & 6.3 \\
Random & 51.0 $\pm$ 0.3 & 52.2 $\pm$ 0.0 & 58.2 $\pm$ 0.2 & 70.1 $\pm$ 0.2 & 58.7 $\pm$ 1.3 & 74.4 $\pm$ 0.5 & 8.0 \\
\bottomrule
\end{tabular}%
}
\end{table}

Table~\ref{tab:results_gretriever_summary} reports hallucination
detection results when \mbox{G-Retriever} is used as the KBQA
pipeline. Unlike KAPING, \mbox{G-Retriever} fine-tunes an LLM on
the KBQA dataset to select answer nodes from the PCST subgraph; the
same subgraphs are shared across both pipelines. SelfCheckGPT is
excluded from this comparison: beyond enabling stochastic sampling
in \mbox{G-Retriever}'s decoding, it would require $N$ additional
\mbox{GPT-5.4-nano} extraction calls per example, which we avoid to
limit API costs. This experiment targets \methodname's robustness to
the choice of KBQA pipeline, not a full rerun of the baseline suite.
\methodname (GINE) and \methodname (GraphTransformer) tie for the
second-best average rank ($3.0$), behind \mbox{GPT-5-mini} which
requires an expensive API call per query. On AUC-PR, \methodname
(GraphTransformer) leads \mbox{GPT-5-mini} on WebQSP ($+2.8$ pp) and
CWQ ($+6.2$ pp).
\mbox{Llama-4-Scout} trails every \methodname
variant in both F1 and AUC-PR on all three datasets. The pattern confirms that \methodname
generalizes beyond the KAPING pipeline.

\subsubsection{Full Results (Llama-4-Scout)}
\label{sec:additional_results-llama_full}

\begin{table}[!htb]
\small
\caption{Full hallucination detection results on WebQSP (F1, Precision, Recall, Accuracy, AUC-PR). Bold and underlined values indicate the best and second-best result per column, respectively. G, RS, and RT denote the retrieved subgraph, reasoning summary, and reasoning triples provided to the LLM judge. Values with $\pm$ denote mean $\pm$ std over 3 runs.}
\label{tab:results_wqsp}
\centering
\begin{tabular}{llllll}
\toprule
\textbf{Method} & \textbf{F1} & \textbf{Precision} & \textbf{Recall} & \textbf{Accuracy} & \textbf{AUC-PR} \\
\midrule
${\methodname\ }_{\text{(GraphTransformer)}}$ & \textbf{82.0 $\pm$ 0.7} & 84.9 $\pm$ 2.2 & 81.3 $\pm$ 2.9 & \textbf{85.9 $\pm$ 0.1} & \textbf{91.4 $\pm$ 0.1} \\
${\methodname\ }_{\text{(GINE)}}$ & \underline{81.2 $\pm$ 0.3} & 81.5 $\pm$ 0.2 & \underline{83.0 $\pm$ 0.3} & \underline{84.4 $\pm$ 0.4} & 90.2 $\pm$ 0.2 \\
${\methodname\ }_{\text{(GAT)}}$ & 79.8 $\pm$ 1.2 & 82.1 $\pm$ 1.3 & 79.8 $\pm$ 1.6 & 83.9 $\pm$ 1.0 & \underline{90.7 $\pm$ 0.2} \\
Most Frequent & 65.9 & 49.1 & \textbf{100.0} & 49.1 & 49.1 \\
${\text{RoBERTa\ }}_{\text{(fine-tuned)}}$ & 63.7 $\pm$ 5.2 & 61.5 $\pm$ 10.7 & 74.7 $\pm$ 8.2 & 72.8 $\pm$ 20.5 & 70.2 $\pm$ 12.8 \\
${\text{SelfCheckGPT\ }}_{\text{(N=10)}}$ & 61.2 & 54.4 & 69.9 & 56.5 & 52.8 \\
${\text{GPT-5-mini\ }}_{\text{(G+RS+RT)}}$ & 59.2 & \underline{85.3} & 45.3 & 69.3 & 65.5 \\
${\text{SelfCheckGPT\ }}_{\text{(N=5)}}$ & 58.9 & 55.3 & 63.1 & 56.8 & 53.0 \\
${\text{GPT-5-mini\ }}_{\text{(G)}}$ & 55.9 & \textbf{85.5} & 41.6 & 67.9 & 64.2 \\
Random & 49.6 $\pm$ 1.0 & 49.4 $\pm$ 0.9 & 49.9 $\pm$ 1.1 & 50.3 $\pm$ 0.9 & 49.2 $\pm$ 0.5 \\
${\text{Llama-4-Scout\ }}_{\text{(G+RS+RT)}}$ & 41.0 & 79.6 & 27.6 & 61.0 & 57.5 \\
${\text{Llama-4-Scout\ }}_{\text{(G)}}$ & 36.7 & 66.7 & 25.3 & 57.1 & 53.5 \\
${\text{Llama-4-Scout\ }}_{\text{(G+RS)}}$ & 36.5 & 73.4 & 24.3 & 58.5 & 55.0 \\
\bottomrule
\end{tabular}
\end{table}

\begin{table}[!htb]
\small
\caption{Full hallucination detection results on CWQ (F1, Precision, Recall, Accuracy, AUC-PR). Bold and underlined values indicate the best and second-best result per column, respectively. G, RS, and RT denote the retrieved subgraph, reasoning summary, and reasoning triples provided to the LLM judge. Values with $\pm$ denote mean $\pm$ std over 3 runs.}
\label{tab:results_cwq}
\centering
\begin{tabular}{llllll}
\toprule
\textbf{Method} & \textbf{F1} & \textbf{Precision} & \textbf{Recall} & \textbf{Accuracy} & \textbf{AUC-PR} \\
\midrule
${\text{RoBERTa\ }}_{\text{(fine-tuned)}}$ & \textbf{87.7 $\pm$ 0.7} & 87.9 $\pm$ 1.4 & 90.2 $\pm$ 2.2 & \textbf{86.1 $\pm$ 0.5} & \textbf{95.5 $\pm$ 0.1} \\
${\methodname\ }_{\text{(GraphTransformer)}}$ & \underline{87.4 $\pm$ 0.2} & 85.8 $\pm$ 0.2 & 90.2 $\pm$ 0.5 & \underline{83.5 $\pm$ 0.2} & \underline{94.2 $\pm$ 0.3} \\
${\methodname\ }_{\text{(GINE)}}$ & 87.1 $\pm$ 0.3 & 84.7 $\pm$ 0.8 & \underline{90.8 $\pm$ 0.8} & 82.8 $\pm$ 0.4 & 93.6 $\pm$ 0.4 \\
${\methodname\ }_{\text{(GAT)}}$ & 86.3 $\pm$ 0.7 & 84.5 $\pm$ 0.3 & 89.6 $\pm$ 1.4 & 81.7 $\pm$ 0.9 & 93.1 $\pm$ 0.2 \\
${\text{GPT-5-mini\ }}_{\text{(G+RS+RT)}}$ & 84.2 & 93.5 & 76.7 & 79.4 & 88.4 \\
${\text{SelfCheckGPT\ }}_{\text{(N=10)}}$ & 84.0 & 81.0 & 87.2 & 76.2 & 79.8 \\
Most Frequent & 83.5 & 71.7 & \textbf{100.0} & 71.7 & 71.7 \\
${\text{GPT-5-mini\ }}_{\text{(G)}}$ & 82.6 & 90.2 & 76.2 & 77.0 & 85.8 \\
${\text{SelfCheckGPT\ }}_{\text{(N=5)}}$ & 82.3 & 82.5 & 82.1 & 74.7 & 80.6 \\
${\text{Llama-4-Scout\ }}_{\text{(G+RS)}}$ & 61.3 & \underline{94.0} & 45.4 & 58.8 & 81.8 \\
${\text{Llama-4-Scout\ }}_{\text{(G+RS+RT)}}$ & 61.2 & \textbf{94.4} & 45.3 & 58.8 & 82.0 \\
${\text{Llama-4-Scout\ }}_{\text{(G)}}$ & 60.2 & 91.6 & 44.8 & 57.5 & 80.7 \\
Random & 58.3 $\pm$ 0.5 & 71.4 $\pm$ 0.3 & 49.3 $\pm$ 0.5 & 49.4 $\pm$ 0.4 & 71.6 $\pm$ 0.1 \\
\bottomrule
\end{tabular}
\end{table}

\begin{table}[!htb]
\small
\caption{Full hallucination detection results on PUGG (F1, Precision, Recall, Accuracy, AUC-PR). Bold and underlined values indicate the best and second-best result per column, respectively. G, RS, and RT denote the retrieved subgraph, reasoning summary, and reasoning triples provided to the LLM judge. Values with $\pm$ denote mean $\pm$ std over 3 runs.}
\label{tab:results_pugg}
\centering
\begin{tabular}{llllll}
\toprule
\textbf{Method} & \textbf{F1} & \textbf{Precision} & \textbf{Recall} & \textbf{Accuracy} & \textbf{AUC-PR} \\
\midrule
${\text{SelfCheckGPT\ }}_{\text{(N=10)}}$ & \textbf{85.5} & 84.9 & 86.0 & \underline{79.0} & 83.1 \\
${\methodname\ }_{\text{(GraphTransformer)}}$ & \underline{84.3 $\pm$ 1.0} & 78.1 $\pm$ 3.4 & \underline{93.5 $\pm$ 4.8} & 78.8 $\pm$ 1.9 & \textbf{90.0 $\pm$ 0.6} \\
${\text{SelfCheckGPT\ }}_{\text{(N=5)}}$ & 83.9 & 86.5 & 81.4 & 77.6 & 83.7 \\
Most Frequent & 83.5 & 71.6 & \textbf{100.0} & 71.6 & 71.6 \\
${\methodname\ }_{\text{(GINE)}}$ & 83.1 $\pm$ 1.0 & 81.9 $\pm$ 1.0 & 86.0 $\pm$ 2.3 & \textbf{79.1 $\pm$ 0.9} & \textbf{90.0 $\pm$ 0.5} \\
${\methodname\ }_{\text{(GAT)}}$ & 82.7 $\pm$ 2.7 & 79.6 $\pm$ 2.0 & 88.1 $\pm$ 7.7 & 77.4 $\pm$ 3.2 & 88.7 $\pm$ 0.7 \\
${\text{GPT-5-mini\ }}_{\text{(G)}}$ & 82.6 & \underline{96.3} & 72.3 & 78.2 & \underline{89.5} \\
${\text{GPT-5-mini\ }}_{\text{(G+RS+RT)}}$ & 81.2 & 95.1 & 70.9 & 76.6 & 88.3 \\
${\text{RoBERTa\ }}_{\text{(fine-tuned)}}$ & 76.7 $\pm$ 2.6 & 75.1 $\pm$ 1.3 & 83.0 $\pm$ 5.5 & 76.3 $\pm$ 0.9 & 86.0 $\pm$ 0.2 \\
${\text{Llama-4-Scout\ }}_{\text{(G)}}$ & 68.8 & 93.5 & 54.4 & 64.7 & 83.5 \\
${\text{Llama-4-Scout\ }}_{\text{(G+RS)}}$ & 61.3 & 96.0 & 45.1 & 59.3 & 82.6 \\
${\text{Llama-4-Scout\ }}_{\text{(G+RS+RT)}}$ & 60.8 & \textbf{96.7} & 44.3 & 59.0 & 82.7 \\
Random & 58.8 $\pm$ 1.8 & 71.8 $\pm$ 1.6 & 49.8 $\pm$ 1.8 & 50.1 $\pm$ 1.9 & 71.7 $\pm$ 0.8 \\
\bottomrule
\end{tabular}
\end{table}

Tables~\ref{tab:results_wqsp}, \ref{tab:results_cwq},
and~\ref{tab:results_pugg} extend the results reported in
Section~\ref{sec:results-hallucination_detection} with the full
precision, recall, accuracy, and AUC-PR breakdown. A consistent pattern
emerges: LLM-based judges are strongly precision-skewed.
\mbox{GPT-5-mini} (G+RS+RT) reaches $85.3\%$ precision on WebQSP but
only $45.3\%$ recall; Llama-4-Scout variants reach up to $96.7\%$
precision on PUGG at recalls near $44\%$.
\methodname achieves a more balanced trade-off: $84.9\%$ precision and
$81.3\%$ recall on WebQSP ($\text{F1}=82.0$), showing that the F1
advantage over LLM judges is driven primarily by higher recall.
\methodname leads in AUC-PR on WebQSP and PUGG ($91.4$ vs.\ RoBERTa's
$70.2$, and $90.0$ vs.\ GPT-5-mini's $89.5$), and is a close second on
CWQ ($94.2$ vs.\ RoBERTa's $95.5$).

\subsection{Hallucination Correction}
\label{sec:additional_results-hallucination_correction}

\begin{table}[!htb]
\small
\caption{KBQA answer quality before (\underline{Initial}) and after (\underline{Refined}) iterative refinement guided by \methodname hallucination flags. $\Delta$ is the score improvement.}
\label{tab:refinement_kbqa_detailed}
\centering
\begin{tabular}{lrrrrrr}
\toprule
\textbf{Dataset} &  & \textbf{F1} & \textbf{Precision} & \textbf{Recall} & \textbf{Accuracy} & \textbf{Exact Match} \\
\midrule
\multirow{3}{*}{\textbf{WebQSP}} & Initial & 59.5 & 60.2 & 63.2 & 68.6 & 42.8 \\
 & Refined & 72.5 & 75.8 & 72.7 & 79.1 & 60.4 \\
 & $\Delta$  & +13.0 & +15.6 & +9.5 & +10.5 & +17.6 \\
 \midrule
\multirow{3}{*}{\textbf{CWQ}} & Initial & 57.2 & 56.8 & 59.9 & 61.3 & 51.1 \\
 & Refined & 71.7 & 72.3 & 72.2 & 73.7 & 68.8 \\
 & $\Delta$ & +14.5 & +15.5 & +12.3 & +12.4 & +17.6 \\
 \midrule
\multirow{3}{*}{\textbf{PUGG}} & Initial & 53.1 & 52.4 & 56.5 & 58.0 & 45.3 \\
 & Refined & 67.2 & 67.2 & 68.9 & 70.2 & 62.3 \\
 & $\Delta$ & +14.0 & +14.8 & +12.4 & +12.2 & +16.9 \\
 \midrule
 & Mean  $\Delta$ & +13.8 & +15.3 & +11.4 & +11.7 & +17.4 \\
\bottomrule
\end{tabular}
\end{table}

Table~\ref{tab:refinement_kbqa_detailed} extends the F1 and Exact
Match results reported in Section~\ref{sec:results-hallucination_correction}
with Precision, Recall, and Accuracy. Gains are consistent across all five metrics and all three
datasets. Precision improves by $14.8$--$15.6$~pp, while recall
improves by $9.5$--$12.4$~pp; precision gains exceeding recall gains
indicate that refinement primarily eliminates incorrect answer nodes
rather than recovering missing ones. 

\section{Additional Statistics}
\label{sec:additional_statistics}

\subsection{Knowledge Graph Subgraphs}
\label{sec:graph_stats}

\begin{table}[!htb]
\small
\caption{Statistics of the retrieved question-specific subgraphs by PCST used as input to \methodname. \underline{Avg./Max \#Nodes} and \underline{Avg./Max \#Triples} report node and edge statistics per subgraph, respectively.}
\label{tab:graph_stats}
\centering
\begin{tabular}{llrrrr}
\toprule
\textbf{Dataset} & \textbf{Split} & \textbf{Avg. \#Nodes} & \textbf{Max \#Nodes} & \textbf{Avg. \#Triples} & \textbf{Max \#Triples} \\
\midrule
\multirow{3}{*}{\textbf{WebQSP}} & train & 15.5 & 307 & 17.4 & 848 \\
 & val & 16.2 & 302 & 20.7 & 1792 \\
 & test & 16.4 & 282 & 19.9 & 1699 \\
\midrule
\multirow{3}{*}{\textbf{CWQ}} & train & 16.5 & 530 & 18.6 & 1783 \\
 & val & 17.2 & 851 & 19.9 & 2691 \\
 & test & 16.3 & 441 & 17.6 & 497 \\
\midrule
\multirow{3}{*}{\textbf{PUGG}} & train & 97.6 & 13614 & 120.5 & 16765 \\
 & val & 72.2 & 13574 & 83.2 & 16702 \\
 & test & 87.8 & 13575 & 108.0 & 16703 \\
\bottomrule
\end{tabular}
\end{table}

\begin{figure*}[!htb]
    \centering
    \includegraphics[width=0.85\textwidth]{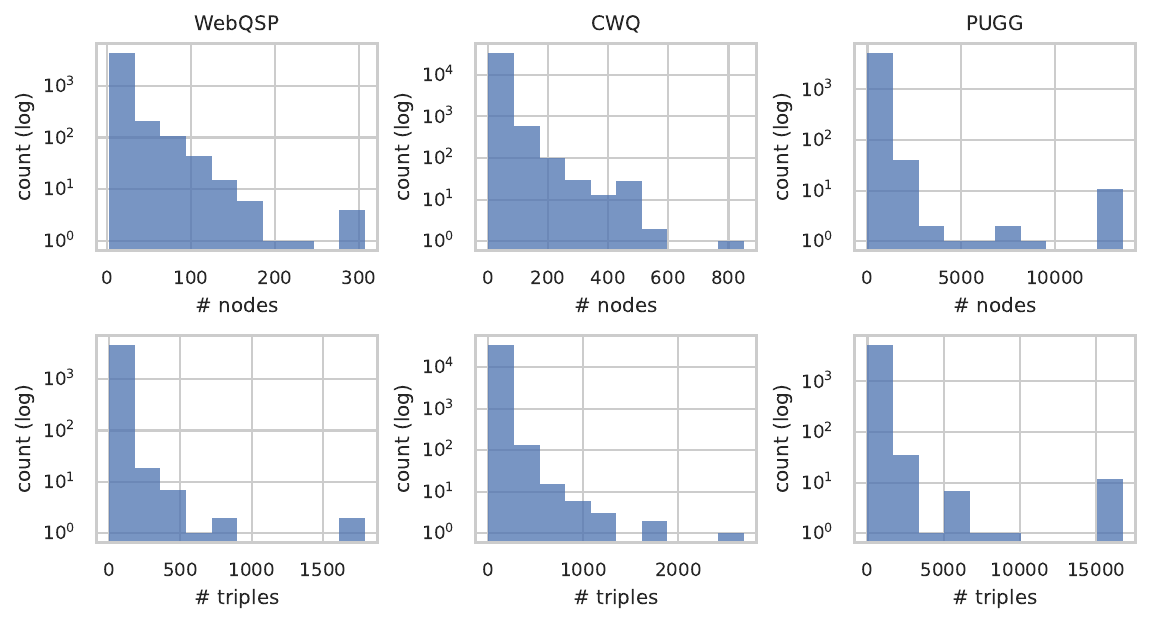}
    \caption{Subgraph size distributions across the three datasets. 
    Top row: number of nodes; bottom row: number of triples. Counts are on a log scale.}
    \label{fig:graph_structure_histograms}
\end{figure*}

Table~\ref{tab:graph_stats} reports subgraph sizes across datasets.
WebQSP and CWQ subgraphs are compact (avg.\ $15$--$17$ nodes,
$17$--$21$ triples), owing to Freebase's sparser entity neighborhood
compared to Wikidata.
PUGG subgraphs are several times larger (avg.\ $72$--$98$ nodes,
$83$--$121$ triples) due to Wikidata's denser structure with many hub nodes, 
even after
filtering nodes connected to more than $1000$ entities.
Maximum subgraph sizes reach thousands of triples across all datasets,
requiring the graph encoder to handle substantial structural variability.
The largest subgraphs may also exceed the context budget used by a
text-based baseline, motivating graph-based verification.
The full subgraph size distributions are shown in
Figure~\ref{fig:graph_structure_histograms}. All three datasets are heavy-tailed.

\subsection{Answer Statistics}
\label{sec:answer_stats}

\begin{table}[!htb]
\small
    \caption{Average answer count per question. \underline{\#GT Ans./Q}: gold (ground truth) answer nodes; \underline{\#LLM Ans./Q}: nodes returned by the LLM-based KBQA method.}
\label{tab:answers_per_ex}
\centering
\begin{tabular}{llrr}
\toprule
\textbf{Dataset} & \textbf{Split} & \textbf{\#GT Ans./Q} & \textbf{\#LLM Ans./Q} \\
\midrule
\multirow{3}{*}{\textbf{WebQSP}} & train & 11.5 & 3.3 \\
 & val & 10.4 & 3.1 \\
 & test & 10.2 & 3.0 \\
\midrule
\multirow{3}{*}{\textbf{CWQ}} & train & 2.2 & 1.6 \\
 & val & 2.3 & 1.7 \\
 & test & 1.9 & 1.7 \\
\midrule
\multirow{3}{*}{\textbf{PUGG}} & train & 1.8 & 1.8 \\
 & val & 1.8 & 1.5 \\
 & test & 1.8 & 1.7 \\
\bottomrule
\end{tabular}
\end{table}

Table~\ref{tab:answers_per_ex} reports the average number of gold and
LLM-predicted answer nodes per question. WebQSP contains substantially
more gold answers per question (avg.\ $10.2$--$11.5$) than CWQ
($1.9$--$2.3$) or PUGG (${\approx}1.8$), reflecting that WebQSP
questions admit multiple valid answer entities. The LLM consistently
under-predicts on WebQSP, returning $3.0$--$3.3$ nodes on average
against $10{+}$ gold answers. On CWQ and PUGG, the LLM returns a similar number of
nodes to the gold count.

\subsection{LLM Abstention and Hallucination Patterns}
\label{sec:abstention_patterns}

\begin{table}[!htb]
\small
\caption{LLM abstention rate (\underline{\%Unknown}) and gold answer coverage (\underline{\%Answer in Graph}: fraction of examples where at least one gold answer node appears in the retrieved subgraph).}
\label{tab:unknown_presence}
\centering
\begin{tabular}{llrr}
\toprule
\textbf{Dataset} & \textbf{Split} & \textbf{\%Unknown} & \textbf{\%Answer in Graph} \\
\midrule
\multirow{3}{*}{\textbf{WebQSP}} & train & 17.6 & 62.9 \\
 & val & 16.2 & 60.6 \\
 & test & 16.8 & 63.0 \\
\midrule
\multirow{3}{*}{\textbf{CWQ}} & train & 37.7 & 41.4 \\
 & val & 34.8 & 42.4 \\
 & test & 37.8 & 39.9 \\
\midrule
\multirow{3}{*}{\textbf{PUGG}} & train & 21.5 & 46.6 \\
 & val & 19.1 & 48.9 \\
 & test & 21.2 & 47.7 \\
\bottomrule
\end{tabular}
\end{table}

\begin{table}[!htb]
\small
\caption{LLM-based KBQA abstention rate (\emph{unknown} answers) conditioned on gold answer presence in the retrieved subgraph. \underline{\%Answer absent}/\underline{\%Answer present}: abstention rate when the gold answer is absent from or present in the subgraph, respectively.}
\label{tab:unknown_by_presence}
\centering
\begin{tabular}{llrr}
\toprule
\textbf{Dataset} & \textbf{Split} & \textbf{\%Answer absent} & \textbf{\%Answer present} \\
\midrule
\multirow{3}{*}{\textbf{WebQSP}} & train & 40.0 & 4.4 \\
 & val & 37.6 & 2.3 \\
 & test & 39.1 & 3.6 \\
\midrule
\multirow{3}{*}{\textbf{CWQ}} & train & 56.6 & 10.9 \\
 & val & 52.5 & 10.8 \\
 & test & 55.4 & 11.2 \\
\midrule
\multirow{3}{*}{\textbf{PUGG}} & train & 37.8 & 2.9 \\
 & val & 35.8 & 1.8 \\
 & test & 39.6 & 1.0 \\
\bottomrule
\end{tabular}
\end{table}

\begin{table}[!htb]
\small
\caption{Hallucination rate among non-abstained LLM-based KBQA answers, conditioned on gold answer presence in the retrieved subgraph. \underline{\%Answer absent}/\underline{\%Answer present}: hallucination rate when the gold answer is absent from or present in the subgraph, respectively. \%Answer absent is always $100\%$ by definition: if the gold answer is not in the subgraph, any non-abstained node returned by the system is hallucinated.}
\label{tab:hallu_by_presence}
\centering
\begin{tabular}{llrr}
\toprule
\textbf{Dataset} & \textbf{Split} & \textbf{\%Answer absent} & \textbf{\%Answer present} \\
\midrule
\multirow{3}{*}{\textbf{WebQSP}} & train & 100.0 & 39.8 \\
 & val & 100.0 & 39.2 \\
 & test & 100.0 & 38.7 \\
\midrule
\multirow{3}{*}{\textbf{CWQ}} & train & 100.0 & 42.5 \\
 & val & 100.0 & 46.9 \\
 & test & 100.0 & 42.4 \\
\midrule
\multirow{3}{*}{\textbf{PUGG}} & train & 100.0 & 41.8 \\
 & val & 100.0 & 45.2 \\
 & test & 100.0 & 42.5 \\
\bottomrule
\end{tabular}
\end{table}

Table~\ref{tab:unknown_presence} shows that a substantial fraction of
LLM responses are abstentions (\emph{unknown}): $16$--$18\%$ on WebQSP,
$19$--$22\%$ on PUGG, and $35$--$38\%$ on CWQ. In all cases, the gold answer 
is present in the retrieved subgraph for
fewer than two-thirds of examples, reflecting the inherent difficulty of
subgraph retrieval. Larger subgraphs increase answer coverage but add
noise and exceed LLM context limits, while smaller subgraphs are more
tractable but risk omitting the correct evidence. 
Tables~\ref{tab:unknown_by_presence} and~\ref{tab:hallu_by_presence}
show how abstention and hallucination rates vary with answer presence.
When the gold answer is absent from the subgraph, the LLM abstains at
rates of $35$--$57\%$ depending on the dataset. When the gold answer is
present, abstention rates drop to $1$--$11\%$ and hallucination among
non-abstaining responses falls to $38$--$47\%$. 

These patterns confirm that hallucination in this setting 
is tightly coupled to whether the
answer evidence is accessible in the retrieved subgraph.
When no gold answer is retrieved, every returned node is labeled
hallucinated by definition. This
motivates both answer verification and improvements to retrieval.

\newpage

\end{document}